\documentclass[11pt]{article}

\usepackage[final]{acl}

\usepackage{times}
\usepackage{latexsym}

\usepackage[T1]{fontenc}

\usepackage[utf8]{inputenc}

\usepackage{microtype}

\usepackage{inconsolata}

\usepackage{graphicx}

\title{Dimple: Discrete Diffusion Parallel Generation for \\Large Multimodal Modal}

\author{\bf Runpeng Yu \quad Xinyin Ma \quad
Xinchao Wang$\thanks{Corresponding Author: Xinchao Wang (Email: xinchao@nus.edu.sg)}$ 
\\
National University of Singapore\\
{\tt\small \{r.yu, maxinyin\}@u.nus.edu} \quad {\tt\small xinchao@nus.edu.sg}
}

\usepackage{subcaption}
\usepackage{booktabs} 
\usepackage{hyperref}

\usepackage{array}

\usepackage[utf8]{inputenc} 
\usepackage[T1]{fontenc}    
\usepackage{hyperref}       
\usepackage{url}            
\usepackage{booktabs}       
\usepackage{amsfonts}       
\usepackage{nicefrac}       
\usepackage{microtype}      
\usepackage{amsmath}
\usepackage{amssymb}
\usepackage{amsthm}
\usepackage{makecell}
\usepackage{graphicx}
\usepackage{multirow}
\usepackage[table,xcdraw]{xcolor}
\definecolor{lightgray}{HTML}{EFEFEF}
\usepackage{subcaption}
\usepackage{caption}
\usepackage{adjustbox}

\usepackage{enumitem}
\usepackage{wrapfig}
\usepackage[utf8]{inputenc}
\usepackage[most]{tcolorbox}
\usepackage[ruled,vlined]{algorithm2e}
\newcolumntype{P}[1]{>{\centering\arraybackslash}p{#1}}
\definecolor{mycolor}{HTML}{abc7df}

\usepackage{amsmath}
\usepackage{amssymb}
\usepackage{mathtools}
\usepackage{amsthm}

\usepackage[capitalize,noabbrev]{cleveref}
\crefname{figure}{Fig.}{Figs.}
\Crefname{figure}{Fig.}{Figs.}
\crefname{table}{Tab.}{Tabs.}
\Crefname{table}{Tab.}{Tabs.}
\crefname{section}{Sec.}{Secs.}
\Crefname{section}{Sec.}{Secs.}

\theoremstyle{plain}

\theoremstyle{definition}

\theoremstyle{remark}

\usepackage[textsize=tiny]{todonotes}

\usepackage[most]{tcolorbox}

\usepackage[most]{tcolorbox} 

\newtcolorbox{takeawaysbox}{
  colback=mycolor!30,
  colframe=mycolor!30!black, 
  boxrule=0.4pt,
  arc=6pt,
  width=\linewidth,
  left=0pt, right=3pt, top=2pt, bottom=2pt,
  enhanced,
  after=\par\smallskip,
}

\newcommand{\takeaways}[1]{%
}

\begin{document}
\maketitle
\begin{abstract}
This paper introduces Dimple and Dimple+, two discrete diffusion multimodal large language models (dMLLMs). Dimple is derived from a discrete diffusion large language model (dLLM) that initially lacks multimodal understanding and acquires this capability through a hybrid training strategy that begins with autoregressive learning and subsequently transitions to discrete diffusion training. In contrast, Dimple+ is initialized from an autoregressive multimodal large language model and attains parallel decoding ability via purely discrete diffusion training. Both models achieve performance on par with their autoregressive counterparts, while Dimple+ further sets a new state of the art among existing dMLLMs. To improve inference efficiency, we propose Confident Decoding, a method that dynamically adapts the number of tokens generated at each iteration. Experimental results demonstrate decoding speedups of $2\times$–$6\times$ with only marginal performance loss. Additionally, we show that the Prefilling technique, previously developed for autoregressive models, can be effectively extended to dMLLMs with bidirectional attention, yielding nearly lossless acceleration of $1.7\times$–$7\times$. Finally, we present the Structure Prior method, which enables fine-grained control over output format and reasoning structure, a capability that is challenging to achieve in autoregressive models.

\end{abstract}    
\section{Introduction}

Recent months have witnessed a surge of interest in applying diffusion models to natural language processing tasks. These Discrete Diffusion Large Language Models (dLLMs) reframe generation as a denoising process, allowing parallel decoding and improved control over output structure. 
The first general framework for diffusion over discrete spaces was introduced in D3PM~\citep{d3pm}, which set up the mathematical foundation of the diffusion langauge model. Subsequently, RDM~\citep{reparam-discrete-diffusion}, MD4~\citep{simplified-masked-diffusion}, and MDLM~\citep{simple-masked-diffusion} reparameterized and simplified the loss for discrete diffusion into a weighted cross-entropy form, which was widely adopted in later researches~\citep{llada,dream,diffullama,llada15}. Other explorations include extending discrete diffusion models to continuous time~\citep{continuoustimedlm}, discrete flow matching~\citep{dfm} and concrete score matching~\citep{concrete_score,sun2023scorebased}. However, these works were typically conducted on models under 2B parameters. Only more recently, LLaDA~\citep{llada}, LLaDA 1.5~\citep{llada15}, Diffu-GPT/LLaMA~\citep{diffullama}, and Dream~\citep{dream} successfully scaled up dLLMs to much larger sizes.

Concurrently, the field of multimodal large language models (MLLMs) has advanced rapidly, with models such as LLaVA~\citep{li2024llavanext-strong}, Qwen-VL~\citep{qwen25vl}, and InternVL~\citep{internvl3} achieving state-of-the-art results on a range of vision-language benchmarks. However, most of the existing MLLMs rely exclusively on autoregressive generation mechanisms.

In this work, we introduce \textbf{Dimple} and \textbf{Dimple+} , two \textit{Discrete Diffusion Multimodal Large Language Models} (dMLLMs). 
Due to the inefficiency of discrete diffusion training in learning multimodal understanding ability, Dimple and Dimple+ utilize a ``first-autoregressive-then-diffusion'' training strategy and ``autoregressive model initialization'' strategy, respectively.
Dimple begins with Dream~\cite{dream}, a dLLM lacking multimodal understanding ability, and first undergoes autoregressive training followed by the  diffusion training. Ablation studies in the supplementary material show that this hybrid approach is significantly more effective for learning multimodal understanding ability than the pure diffusion training. In contrast, Dimple+ is initialized from Qwen2.5-VL~\citep{qwen25vl}, an MLLM with strong multimodal understanding ability but without discrete diffusion-based parallel decoding ability. By applying only discrete diffusion training, Dimple+ acquires the parallel decoding capability while inheriting Qwen2.5-VL’s strong multimodal understanding ability, achieving the SOTA performance among dMLLMs.

To further improve inference efficiency and controllability, we propose \textbf{Confident Decoding}, which dynamically adjusts the number of tokens decoded per iteration based on a confidence threshold. This technique accelerates decoding by approximately 2× to 6× with only minor performance degradation. We also demonstrate the feasibility of applying the \textbf{Prefilling} technique in autoregressive models to dMLLMs with bidirectional attention. Prefilling achieves nearly lossless speedups of 1.7× to 7× on most benchmarks. In addition, we explore the \textbf{Structure Prior} technique, which enables fine-grained control over response structure. With Structure Prior, dMLLMs can perform early answer prediction, structured reasoning, and in-place output formatting, which are difficult to achieve with autoregressive models.

\paragraph{Contributions.}Our key contributions are as follows:
\begin{itemize}[leftmargin=20pt, itemsep=0pt, parsep=0pt]
    \item We introduce \textbf{Dimple} and \textbf{Dimple+}, both of which achieve performance comparable to their autoregressive baselines under the same training budget. Moreover, Dimple+ establishes state-of-the-art results among dMLLMs.
    \item We propose \textbf{Confident Decoding}, re-implement \textbf{Prefilling}, and explore \textbf{Structure Prior} for dMLLM inference that improve the efficiency, flexibility, and controllability.
\end{itemize}

\section{Preliminary}

\paragraph{Discrete Diffusion Large Language Models (dLLMs).} dLLMs are a class of generative models that conceptualize text generation as a denoising process over discrete time steps. Let $x_0 \sim p_{\text{data}}(x_0)$ denote the original token sequence, and $x_t$ its noisy version at time $t \in [0, T]$. The forward noising process is defined as a Markov chain $q(x_{1:T} | x_0) = \prod_{t=1}^T q(x_t | x_{t-1})$, progressively adding noise to $x_0$. For discrete token modeling with an absorbing state, this is formalized as:
\vspace{-0.5em}
\begin{align}
q(x_t | x_{t-1}) &= \text{Cat}(x_t; Q_t^\top x_{t-1}), \nonumber\\
q(x_t | x_0) &= \alpha_t x_0 + (1 - \alpha_t) \mathbf{m}, \nonumber\\
\alpha_t &= \prod_{i=1}^t (1 - \beta_i),
\end{align}
where $Q_t = (1 - \beta_t)I + \beta_t \mathbf{1} \mathbf{m}^\top$, with $\mathbf{m}$ being the one-hot representation of a special \texttt{[MASK]} token.
The reverse (generative) process is modeled as:
\begin{align}
p_\theta(x_{0:T}) &= p_\theta(x_T) \prod_{t=1}^T p_\theta(x_{t-1} | x_t) \nonumber\\
&= \prod_{t=1}^T q(x_{t-1} | x_0) p_\theta(x_0 | x_t),
\end{align}
where each transition $p_\theta(x_{t-1} | x_t)$ is learned to approximate the reverse of the noising process with bidirectional attention. 
The training loss of Discrete Diffusion, derived from the variational lower bound of $p(x_0)$, is a weighted cross-entropy loss:
\begin{align}
\mathcal{L}_{D}&=\mathbb{E}_{t}[\mathcal{L}_t],\nonumber\\
\mathcal{L}_t&=\mathbb{E}_{q(x_t|x_0)}\left[-\sum_{n=1}^N\frac{1}{t}\delta_{x_t^n,\mathbf{m}}(x_0^n)\ log[f_\theta(x_t)^n]\right].\label{eq:diffusion_loss}
\end{align}
This loss can be implemented as a Masked Language Modeling loss with a random mask ratio. The mask ratio is function of the time index $t$, which is a random variable. The expectation $\mathbb{E}_{t}$ is taken over a uniform distribution of time steps $t\in[0,1]$. In practice, for each training instance, one time index $t$ is samples to determine the proportion of tokens to be masked. The inner expectation $\mathbb{E}_{q(x_t|x_0)}$ is taken over the forward transition probability defined in Eq. 1. $N$ denotes the number of tokens in the current input sample $x_0$. Inside the summation $\sum_{n=1}^N$, a weighted cross-entropy is computed over each token. The weights consist of two parts: (1) a time-dependent factor $\frac{1}{t}$, and (2) an indicator function $\delta_{x_t^n, \mathbf{m}}$ which ensures that the loss is only computed for masked tokens.  Finally, $f_\theta$ denotes the model's prediction.

\paragraph{Challenges of Discrete Diffusion Training.} 
Due to the use of bidirectional attention and the limited number of sampled timesteps, training with Discrete Diffusion directly incurs higher computational costs compared to autoregressive (AR) training~\citep{prabhudesai2025diffusionbeatsautoregressivedataconstrained,yu2025discretediffusionlargelanguage}.
\begin{enumerate}[leftmargin=20pt, itemsep=0pt, parsep=0pt]
    \item Compared to next-token prediction loss, masked language modeling has lower utilization of the training corpus.  In Auto-Regressvie training, every response token receives supervision. In contrast, in discrete diffusion training, only tokens where $\delta_{x_t^n, \mathbf{m}} = 1$ (i.e., masked tokens) are used to compute the loss. Take linear masking scheduling as an example, with $t \sim \text{Uniform}(0, 1)$, only about half the tokens are supervised.
    \item In diffusion training, the supervision signal does not cover the entire generation trajectory. Discrete diffusion inference proceeds iteratively as $x_T \rightarrow x_{T-1} \rightarrow \cdots \rightarrow x_0$, but during training each sample is trained on only a single, or at most a few, randomly sampled steps: $x_t \rightarrow x_{t-1}$. Consequently, it is impossible to cover the complete generation trajectory for each sample. In contrast, in autoregressive training, next-token prediction with causal attention ensures that the supervision signal covers every step of the generation process for each sample.
\end{enumerate}
These two challenges result in discrete diffusion models performing worse than autoregressive models \textbf{under the same training budget}. Empirical evidence is presented in \cref{sec:exp}. 

\vspace{-0.5em}
\section{Methodology}
\vspace{-0.5em}
\subsection{Training}
We adopt two strategies to mitigate the limitations of discrete diffusion training. The first applies autoregressive training prior to discrete diffusion training to enhance the model’s learning of  multimodal understanding ability. The second initializes the model with the weights of an autoregressive MLLM, followed by discrete diffusion training to enable the acquisition of parallel decoding capabilities.
Details of the training configuration, masking strategies, and sample padding are provided in the supplementary material.

\paragraph{Strategy I (Dimple): Starting from a Diffusion Large Language Model}
We denote the model obtained using Strategy I as Dimple.
We initialize the model with Qwen2.5-VL’s vision encoder~\citep{qwen25vl}, DLM Dream~\citep{dream}, and a randomly initialized two-layer projector~\citep{llava15}. Following the alignment and instruction-tuning paradigm of LLaVA~\citep{llava}, we design a three-stage training pipeline: Autoregressive Alignment (AA), Autoregressive Instruction Tuning (AT), and Diffusion Instruction Tuning (DT).

In AA and AT, we modify Dream’s attention mechanism by replacing the full attention mask with a causal attention mask and applying a next-token prediction loss. We adopt the LLaVA-NEXT~\citep{liu2024llavanext} training recipe and its datasets. During AA, only the projector is trainable; during AT, all components except the vision encoder are trainable. In the experiments reported in \cref{sec:exp}, we refer to the model obtained after AA as the aligned model.
In the DT phase, we transition back to the diffusion-based generation framework. The full attention mask in Dream is restored, and the loss defined in \cref{eq:diffusion_loss} is applied. The LLaVA-NEXT instruction-following dataset is reused for this phase. All components except the vision encoder are trainable.

\paragraph{Strategy II (Dimple+): Starting from a Autoregressive Multimodal Large Language Model}
We denote the model obtained using Strategy II as Dimple+.
For initialization, we adopt the full Qwen2.5-VL model~\citep{qwen25vl}. Training consists of a single Diffusion Instruction Tuning (DT) phase. Specifically, we modify the attention mask in Qwen2.5-VL to full attention and apply the loss defined in \cref{eq:diffusion_loss}, updating all components except the vision encoder. The training data consists of the instruction-following dataset from LLaVA-NEXT~\citep{liu2024llavanext} and 1.3M purely text-based instruction-following samples collected from the MAmmoTH-VL~\citep{guo2025mammothvlelicitingmultimodalreasoning} dataset.
 Since Qwen2.5-VL~\citep{qwen25vl} has already undergone autoregressive alignment and autoregressive instruction tuning, this implementation actually delegates the AA and AT phases to Qwen2.5-VL.

\subsection{Inference Techniques}

\paragraph{Review: Confidence-Based Decoding.}
In discrete diffusion models, the generation process proceeds iteratively over a fixed number of steps. At each step, a subset of masked positions is selected and updated based on their confidence scores. 
Let \( \mathbf{x}_t \) denote the input sequence at step \( t \). Let \( \mathbf{z}_t \in \mathbb{R}^{L \times V} \) be the corresponding logits, where \( V \) is the vocabulary size. Taking MaskGIT's decoding algorithm~\citep{maskgit} as an example, the decoding process is as follows.
\begin{align*}
    &\text{Probabilities: }\mathbf{p}_t = \text{softmax}(\mathbf{z}_t / \tau); \\
    &\text{Confidence: }\quad c_t^{(i)} = \max( p_t^{(i)});\\
    &\text{Select \( K \) positions: } \mathcal{I}_t = \text{TopK}(c_t, K); \\
    &\text{Sample tokens } x^{(i)}_t \sim \text{Categorical}(p_t^{(i)}) \text{ for }  i \in \mathcal{I}_t.
\end{align*}
This approach ensures that tokens with higher prediction confidence are updated earlier.

\paragraph{Confident Decoding.}
In previous confidence-based decoding~\citep{maskgit,dream,llada}, the number of tokens decoded per step is fixed. However, we argue that decoding should adapt to the semantic structure of the text: some steps may allow many tokens to be confidently predicted, while others may necessitate more caution. We therefore propose \textbf{Confident Decoding}, which dynamically adjusts the number of tokens updated per step based on a fixed confidence threshold \( \gamma \in (0, 1) \). Formally, at each step \( t \), we define $\mathcal{I}_t = \{i \mid c_t^{(i)} \geq \gamma\}$, 
where \( c_t^{(i)} \) is the confidence score for position \( i \). If  $\mathcal{I}_t$ is non-empty, all tokens at positions \( \mathcal{I}_t \) are decoded. Otherwise, we decode the token at the position with the highest confidence. This method enables:
\begin{itemize}[leftmargin=20pt, itemsep=0pt, parsep=0pt, topsep=0pt]
    \item decoding multiple tokens simultaneously when model is highly confident, improving efficiency;
    \item avoiding low-confidence updates, preserving generation quality.
\end{itemize}
This adaptive mechanism reduces unnecessary updates of uncertain tokens and enhances both quality and speed of generation.

\begin{table*}[t]
\centering 
\small 
\makebox[\textwidth]{\resizebox{1\textwidth}{!}{
\begin{tabular}{@{}l>{\columncolor{mycolor!50}}P{2.3cm}@{$\ \ $}>{\columncolor{mycolor!50}}P{2cm}@{$\ \ $}>{\columncolor{mycolor!50}}P{2cm}@{$\ \ $}>{\columncolor{mycolor!50}}P{2cm}@{$\ \ $}c@{$\ \ $}c@{}}
\toprule
Model                     & \makecell{Dimple-7B\\(ours)} & \makecell{Dimple-7B\\(AR Baseline)} & \makecell{Dimple-7B+\\(ours)} & \makecell{Dimple-7B+\\(AR Baseline)} & LLaVA-NEXT-7B & Qwen2.5-VL-7B \\ \midrule
\#Training Samples/Tokens & 1.3M/0.8B        & 1.3M/0.8B               & 2.6M/1.6B         & 2.6M/1.6B                & 1.3M/-        & -/2.6T       \\ \midrule
Base LLM/MLLM             & Dream(Qwen2.5)   & Qwen2.5                 & Qwen2.5 VL        & Qwen2.5 VL               & Vicuna-1.5    & Qwen2.5       \\ \midrule
GQA                       & 59.24            & 59.1                    & 63.2              & \textbf{64.1}            & 64.8          & 60.0          \\
MMBench\_en\_test         & 74.6             & 73.9                    & \textbf{81.5}     & 81.2                     & 68.7          & 83.5          \\
MME\_percpetion           & 1514             & 1477                    & \textbf{1558}     & 1535                     & 1519          & 1533          \\
MME\_cognition            & 432              & 457                     & \textbf{627}      & 610                      & 332           & 620           \\
MME\_total                & 1946             & 1934                    & \textbf{2185}     & 2145                     & 1851          & 2153          \\
POPE                      & 86.2             & 86.3                    & \textbf{87.7}     & 86.9                     & 86.7          & 87.9          \\
MMMU\_val                 & 45.2             & 45                      & \textbf{50.2}     & 49.1                     & 35.8          & 58.6          \\
SQA\_img                  & 77.1             & 77.2                    & 79.1              & \textbf{83.6}            & 72.8          & 52.1          \\
AI2D                      & 74.4             & 73.6                    & \textbf{82.4}     & 80                       & 65.4          & 83.9          \\
ChartQA                   & 63.4             & 58.5                    & 74.7              & \textbf{74.8}            & 54.9          & 87.3          \\
TextQA\_eval              & 61.6             & 53.6                    & \textbf{65.4}      & 64.1            & 64.8          & 77.8          \\
OCRBench                  & 565              & 486                     & \textbf{699}      & 674                      & 490           & 783           \\
MathVista\_test\_mini     & 42.3             & 42.6                    & \textbf{61.0}       & 59.0                       & 33.0          & 68.2          \\
MMVet                     & 41.2             & 37.8                    & 47.5              & \textbf{49.2}            & 47.3          & 67.1          \\
Average                   & 62.4             & 60.6                    & \textbf{70.7}     & 70.2                     & 58.5          & 73.8          \\\bottomrule
\end{tabular}}}
\vspace{-0.5em}
\caption{Benchmark Performance Comparison. Dimple outperforms autoregressive baselines trained on similar data scales. In the table, the highlighted values in each row indicate the best-performing results among the following four models: Dimple, Dimple (AR Baseline), Dimple+, and Dimple+ (AR Baseline).
}
\vspace{-0.5em}
\label{tab:main}
\end{table*}
\begin{table*}[]
\makebox[\textwidth]{\resizebox{0.8\textwidth}{!}{
\begin{tabular}{lllllllll}
\toprule
Model          & MME\_p & MME\_c & MME total & MMMU & AI2D & ChartQA & TextQA & MathVista \\ \midrule
\rowcolor{mycolor!50}Dimple+ (ours) & \textbf{1558}           & \textbf{627}           & \textbf{2185}      & \textbf{58.2} & \textbf{82.4} & 76.7    & \textbf{65.4}   & \textbf{61.0}        \\
LaViDA-L       & 1366           & 341           & 1707      & 43.3 & 70.0   & 64.6    & 56.3   & 44.8      \\
LaViDA-D       & 1463           & 389           & 1852      & 42.6 & 59.0   & 61.0      & 57.1   & 42.1      \\
LLaDA-V        & 1507           & 491           & 1998      & 48.6 & 77.8 & \textbf{78.3}    & 65.3   & 59.7      \\ \bottomrule
\end{tabular}}}
\vspace{-0.5em}
\caption{Performance comparison with other Discrete Diffusion MLLMs. Our Dimple+ outperforms other models with a clear Margin on most of the benchmarks. In the table, MME\_p, MME\_c, and MathVista are abbreviations for MME\_perception, MME\_coginition, and MathVista\_test\_mini, respectively. 
}
\vspace{-0.5em}
\label{tab:compare_dmllm}
\end{table*}

\paragraph{Prefilling.}
Let \( L_{\text{prompt}} \) denote the lengths of the prompt, which includes the question, image, and system prompts. Let \( L_{\text{answer}} \) denote the length of the response. With vision tokens, the prompt can become unignorably long.
The use of a full attention mask in diffusion decoding results in a quadratic complexity $\mathcal{O}((L_{\text{prompt}} + L_{\text{answer}})^2)$
per decoding step.
To alleviate this cost, we re-implement the \textbf{Prefilling} strategy in the autoregressive models, which saves the key-value pairs of the prompt tokens after the first generation step and reuse them in the following steps, reducing the complexity to $\mathcal{O}(L_{\text{answer}}^2)$. But, due to the use of a full attention mask in dMLLM, the prefilling technique is not strictly lossless.


\section{Experiments}
All training configurations, evaluation metrics, generation hyperparameters used during evaluation, and other experimental details are provided in the supplementary material. The sensitivity analysis of the generation hyperparameters and additional ablation studies on the ``AR-then-Diffusion'' training strategy are also included in the supplementary material.

\label{sec:exp}
\subsection{Benchmark Performance}
\paragraph{Comparison with AR Baseline.} We compare the performance of Dimple and Dimple+ with their corresponding AR baselines in \cref{tab:main}. Each AR baseline is trained with the same data, parameter initialization, and total number of training iterations as Dimple and Dimple+, ensuring a fair comparison. 
As shown by the results, both Dimple and Dimple+ outperform their AR baselines. Dimple achieves an average score of 62.4\%, 1.8\% higher than its baseline, while Dimple+ surpasses the baseline on 9 of 12 benchmarks. These results indicate that, under the same training budget, dMLLM can match the performance of autoregressive models on instruction-following and visual understanding tasks. 
We also include the performance of LLaVA-NEXT~\citep{liu2024llavanext} and Qwen2.5-VL~\citep{qwen25vl} as references. Compared with Qwen2.5-VL, Dimple+ and its AR baseline improve on some benchmarks but regress on others. On the one hand, this results from insufficient task coverage in the training data, which improves performance on some benchmarks but degrades it on others. On the other hand, it also suggests that learning for parallel decoding is not entirely orthogonal to multimodal understanding; instead, the process of learning parallel decoding affects multimodal understanding ability.
\takeaways{dMLLMs like Dimple can achieve performance comparable to their AR baselines, under the same training budget.
}

\paragraph{Comparison with Other dMLLMs.} We compared the performance of Dimple+, LaViDA~\citep{lavida}, and LLaDA-V~\citep{lladav} in \cref{tab:compare_dmllm}. Dimple+ significantly outperforms the other two baselines on most benchmarks, achieving state-of-the-art results for dMLLM models. Moreover, training Dimple+ requires substantially lower computational cost. Specifically, Dimple+ is trained on only 2.6M samples, whereas training LLaDA-V involves more than 17M samples.
This demonstrates the effectiveness of the ``Autoregressive Model Initialization'' strategy for dMLLMs. 
\takeaways{Initializing with a strong AR-MLLM can substantially enhance the performance of dMLLMs.}

\subsection{Ablation on Prefilling}

\begin{table*}[t]
\centering 
\small 
\renewcommand{\arraystretch}{1.3} 
\makebox[\textwidth]{\resizebox{1\textwidth}{!}{
\begin{tabular}{@{}cccccccccc}
\toprule
R.L.                & Prefilling                                  & Metric                      & B.S.                            & SQA                          & MMMU                         & MME                          & POPE                         & ChartQA                      & AI2D                         \\ \midrule
                    &                                             & \cellcolor{mycolor!50}ACC & \cellcolor{mycolor!50}1 or 32 & \cellcolor{mycolor!50}84.4 & \cellcolor{mycolor!50}43.4 & \cellcolor{mycolor!50}1946 & \cellcolor{mycolor!50}85.0 & \cellcolor{mycolor!50}51.7 & \cellcolor{mycolor!50}74.5 \\
                    &                                             &                             & 1                               & 34.8                         & 21.3                         & 22.7                         & 22.4                         & 21.5                         & 22.9                         \\
                    & \multirow{-3}{*}{×}                         & \multirow{-2}{*}{TPS}       & 32                              & 45.9                         & 35.5                         & 37.0                         & 36.9                         & 36.3                         & 37.2                         \\ \cmidrule(l){2-10} 
                    &                                             & \cellcolor{mycolor!50}ACC & \cellcolor{mycolor!50}1 or 32 & \cellcolor{mycolor!50}84.4 & \cellcolor{mycolor!50}43.4 & \cellcolor{mycolor!50}1822 & \cellcolor{mycolor!50}85.1 & \cellcolor{mycolor!50}51.3 & \cellcolor{mycolor!50}74.4 \\
                    &                                             &                             & 1                               & 49.3 \underline{($\times$1.32)}         & 19.9 \underline{($\times$1.67)}         & 21.3 \underline{($\times$1.63)}         & 20.7 \underline{($\times$1.65)}         & 20.1 \underline{($\times$1.69)}         & 21.5 \underline{($\times$1.62)}         \\
\multirow{-6}{*}{4} & \multirow{-3}{*}{\checkmark} & \multirow{-2}{*}{TPS}       & 32                              & 168.5 \underline{($\times$3.42)}        & 62.8 \underline{($\times$3.16)}         & 77.6 \underline{($\times$3.64)}         & 78.2 \underline{($\times$3.78)}         & 74.4 \underline{($\times$3.70)}          & 79.7 \underline{($\times$3.71)}         \\ \midrule
                    &                                             & \cellcolor{mycolor!50}ACC & \cellcolor{mycolor!50}1 or 32 & \cellcolor{mycolor!50}84.1 & \cellcolor{mycolor!50}45.3 & \cellcolor{mycolor!50}1900 & \cellcolor{mycolor!50}84.8 & \cellcolor{mycolor!50}58.4 & \cellcolor{mycolor!50}74.5 \\
                    &                                             &                             & 1                               & 29.2                         & 19.3                         & 19.3                         & 19.5                         & 19.1                         & 19.7                         \\
                    & \multirow{-3}{*}{×}                         & \multirow{-2}{*}{TPS}       & 32                              & 44.9                         & 39.7                         & 40.6                         & 40.4                         & 40.4                         & 40.7                         \\ \cmidrule(l){2-10} 
                    &                                             & \cellcolor{mycolor!50}ACC & \cellcolor{mycolor!50}1 or 32 & \cellcolor{mycolor!50}84.0 & \cellcolor{mycolor!50}45.2 & \cellcolor{mycolor!50}1905 & \cellcolor{mycolor!50}84.0 & \cellcolor{mycolor!50}58.4 & \cellcolor{mycolor!50}74.2 \\
                    &                                             &                             & 1                               & 49.2 \underline{($\times$1.54)}         & 18.1 \underline{($\times$2.06)}         & 21.4 \underline{($\times$2.10)}          & 20.7 \underline{($\times$2.07)}         & 20.0 \underline{($\times$2.12)}           & 21.6 \underline{($\times$2.07)}         \\
\multirow{-6}{*}{8} & \multirow{-3}{*}{\checkmark} & \multirow{-2}{*}{TPS}       & 32                              & 288.2 \underline{($\times$5.86)}        & 116.5 \underline{($\times$6.44)}        & 145.1 \underline{($\times$6.80)}         & 147.4 \underline{($\times$7.12)}        & 138.9 \underline{($\times$6.95)}        & 147.9 \underline{($\times$6.85)}        \\ \bottomrule
\end{tabular}}}
\vspace{0.5em}
\caption{Ablation Study on the Prefilling Technique. We compared model performance (ACC) and inference speed (Tokens Per Second, TPS) with and without the use of prefilling. Performance-related rows are highlighted in \colorbox{mycolor!50}{blue}. The underlined values indicate the speedup factor achieved with prefilling. ``R.L.'' and ``B.S.'' denote Response Length and Batch Size, respectively. The results demonstrate that although, in theory, prefilling is not lossless in dMLLMs, it  causes minor drops in performance on most benchmarks while delivering substantial improvements in inference speed.
}
\vspace{-2em}
\label{tab:prefill}
\end{table*}
The following experiments investigate the impact of the prefilling technique on both performance and generation speed in dMLLMs. We conducted experiments with response lengths of 4 and 8, evaluating benchmark performance with and without prefilling. As shown in \cref{tab:prefill}, prefilling does not lead to significant performance degradation on most datasets, with the average performance drop across all benchmarks being only 0.8\%. This indicates that, in the current model, visual perception and the utilization of image tokens remain largely unchanged during text generation.
Next, we examined the effect of prefilling on generation speed. By adjusting the batch size to 1 and 32, we simulated scenarios of low and high GPU utilization, respectively. The results show that: (1) under low GPU utilization (batch size = 1), prefilling yields an average speedup of 1.79×; (2) under high GPU utilization (batch size = 32), the acceleration effect is more pronounced, reaching up to 7×.
\takeaways{Prefilling provides dMLLM with an almost lossless speedup on most benchmarks, up to 7×, by avoiding redundant computation of input key-value states.
}

\subsection{Ablation on Confident Decoding}

We evaluate the Confident Decoding from two perspectives: \textit{efficiency} and \textit{performance}. Confident Decoding is designed to reduce the number of inference steps, thereby accelerating generation. The last row in \cref{tab:confident_decoding} shows the actual number of decoding steps $k$ under a fixed response length of 8, along with the corresponding speedup factors (in parentheses).
As shown, Confident Decoding reduces $k$ to 1.25~2.87, resulting in approximately 3× to 6× speedup. 

Next, we compare Confident Decoding with previous decoding strategies~\citep{dream,llada}: MaskGIT, Entropy, and Random. Unlike these methods, Confident Decoding is the first algorithm capable of dynamically adjusting the number of decoding steps, whereas the others require a predefined fixed value of $k$. For fairness, we set the number of decoding steps for the other strategies to either the response length (8) or one-quarter of it (2).
As shown in \cref{tab:confident_decoding}, reducing the decoding steps to 2 leads to a substantial performance drop for other methods—up to 10\% on CharQA for MaskGIT and Entropy. In contrast, Confident Decoding achieves performance comparable to the 8-step baselines with only about 2 steps. These results demonstrate that Confident Decoding enables faster inference without sacrificing accuracy.

\takeaways{Confident Decoding effectively accelerates decoding by up to 6× through reducing the number of generation steps, while incurring only minimal performance loss.
}

\begin{table}[]
\centering
\small
\makebox[\linewidth]{\resizebox{1\linewidth}{!}{
\begin{tabular}{lccccc}
\toprule
                                    & \textit{Steps}              & CharQA      & SQA         & POPE        & AI2D        \\ \midrule
\multirow{2}{*}{MaskGit}            & 8                  & 63.0        & 77.2        & 86.1        & 72.9        \\
                                    & 2                  & 52.2        & 76.2        & 85.5        & 71.1        \\
\multirow{2}{*}{Entropy}            & 8                  & 63.0        & 77.2        & 86.0        & 73.0        \\
                                    & 2                  & 54.9        & 76.5        & 85.1        & 72.9        \\
\multirow{2}{*}{Random}             & 8                  & 59.3        & 76.1        & 86.0        & 72.8        \\
                                    & 2                  & 57.1        & 76.0        & 85.2        & 71.0        \\\midrule
\rowcolor{mycolor!50}\makecell{Confident\\Deocding} & \textit{Acc} & 62.5        & 77.1        & 86.4        & 73.1        \\
\rowcolor{mycolor!50}                                    &  \textit{Steps}                  & 2.87 (x2.8) & 1.41 (x5.7) & 1.25 (x6.4) & 1.46 (x5.5) \\ \bottomrule
\end{tabular}
}}
\vspace{0.5em}
\caption{Comparison of confident decoding with other decoding algorithms. Confident decoding significantly reduces the number of decoding steps while maintaining nearly the same performance, thereby improving decoding efficiency.
}
\vspace{-1em}
\label{tab:confident_decoding}
\end{table}

\section{Generation Behaviors}

\begin{table*}[t]
\centering
\begin{tcolorbox}[
  width=\linewidth,
  boxrule=3pt,
  colframe=lightgray,
  colback=lightgray!35,
  boxsep=1pt,
  left=3pt,
  right=3pt,
  top=1pt,
  bottom=1pt
]
\begin{minipage}{\textwidth}
\flushleft
\small
\begin{minipage}[b]{0.6\textwidth}
\textbf{Question}: What is the common item in the two images?\\
\textbf{Response}: In the first image, there  is a pair of black scissors. 
In the second image, there  is a pair of black scissors and a pink plate with a pig face on it. 
The common item in the two images  is the scissors.\\
\textbf{Response Length / \#Remaining Tokens / Actual Iterations}:\\  \centerline{64/42/30}\\
\textbf{Generation History}:
\vspace{-2em}
\end{minipage}
\hfill
\begin{minipage}[t]{0.37\textwidth}
\includegraphics[height=1.9cm]{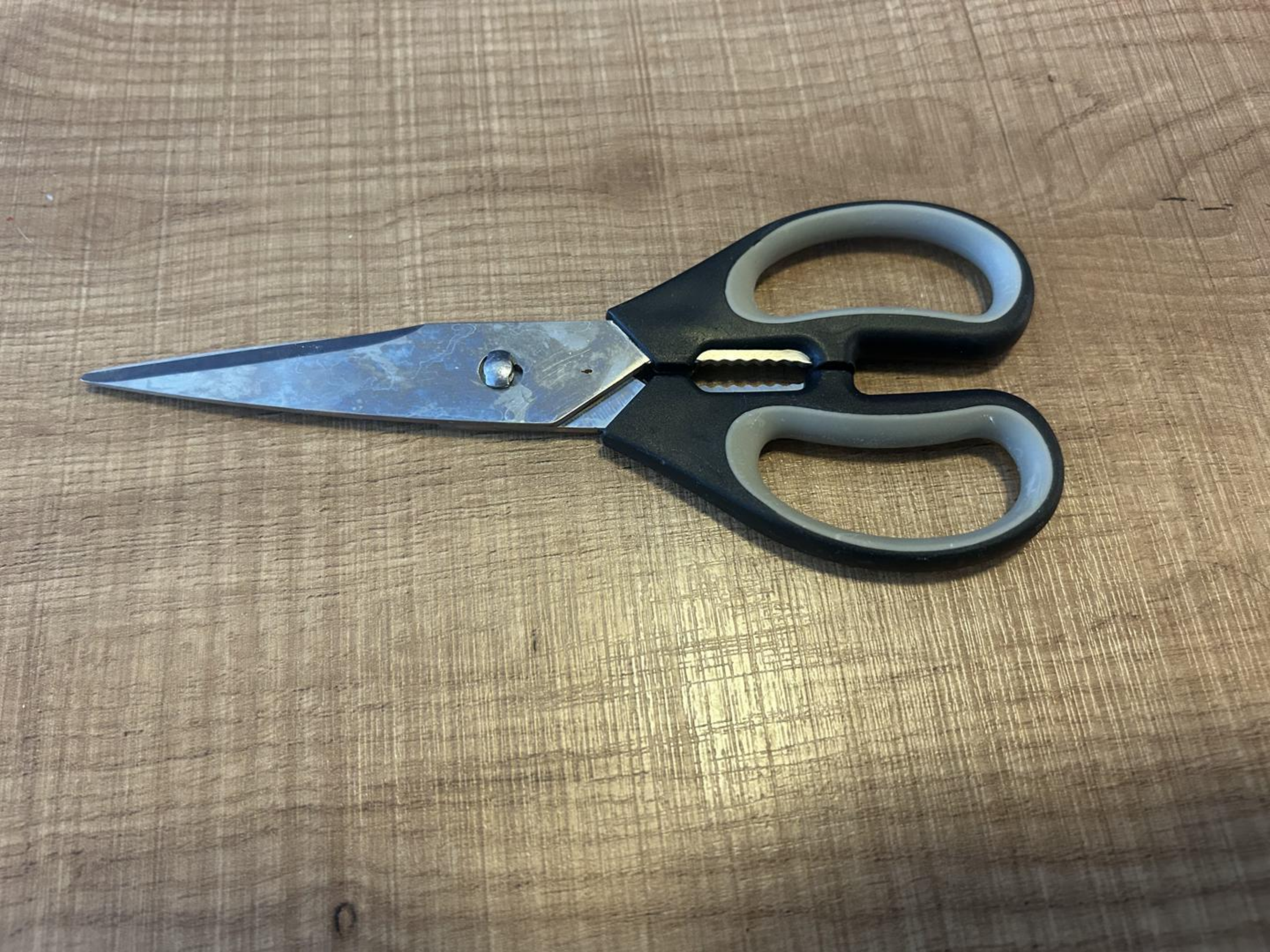} \includegraphics[height=1.9cm]{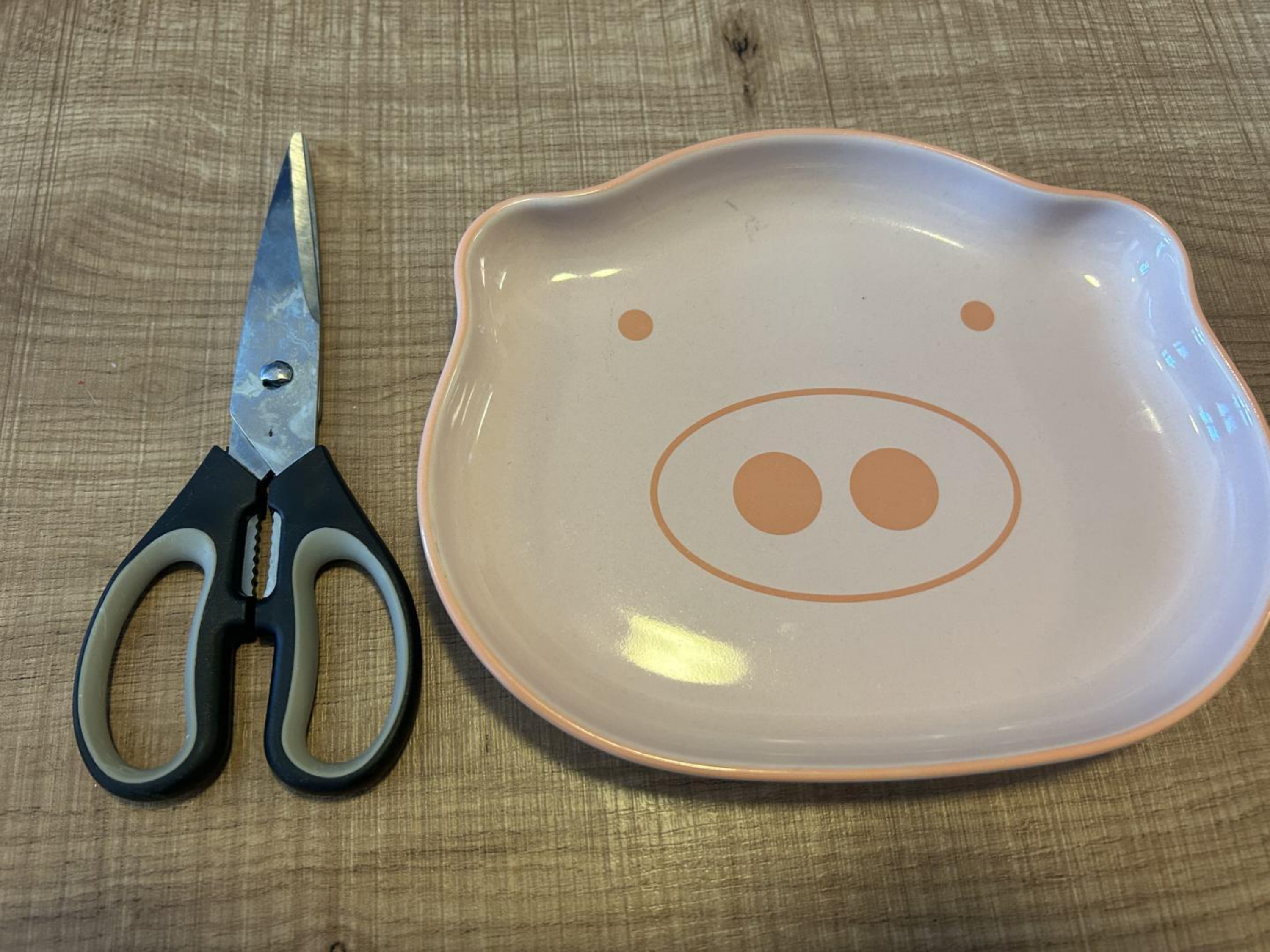}\\\\
\end{minipage}
\makebox[\textwidth]{\resizebox{\textwidth}{!}{\begin{tabular}{ccccccccccccccccc}\midrule
\textcolor[HTML]{1F77B4}{In} & \textcolor[HTML]{1F77B4}{$\ $the} & \textcolor[HTML]{1F77B4}{$\ $first} & \textcolor[HTML]{1F77B4}{$\ $image} & \textcolor[HTML]{1F77B4}{,} & \textcolor[HTML]{1F77B4}{$\ $there} & \textcolor[HTML]{1F77B4}{$\ $} & \textcolor[HTML]{326EA5}{$\ $is} & \textcolor[HTML]{2B71AA}{$\ $a} & \textcolor[HTML]{3F689B}{$\ $pair} & \textcolor[HTML]{456696}{$\ $of} & \textcolor[HTML]{8E465E}{$\ $black} & \textcolor[HTML]{456696}{$\ $scissors} & \textcolor[HTML]{386BA0}{.} & \textcolor[HTML]{3F689B}{$\ $
} & \textcolor[HTML]{1F77B4}{In} & \textcolor[HTML]{1F77B4}{$\ $the} \\
- & - & - & - & - & - & - & 2 & 1 & 4 & 5 & 16 & 5 & 3 & 4 & - & - \\\midrule
\end{tabular}}}

\makebox[\textwidth]{\resizebox{\textwidth}{!}{\begin{tabular}{ccccccccccccccc}
\textcolor[HTML]{1F77B4}{$\ $second} & \textcolor[HTML]{1F77B4}{$\ $image} & \textcolor[HTML]{1F77B4}{,} & \textcolor[HTML]{1F77B4}{$\ $there} & \textcolor[HTML]{1F77B4}{$\ $} & \textcolor[HTML]{326EA5}{$\ $is} & \textcolor[HTML]{326EA5}{$\ $a} & \textcolor[HTML]{7A4E6D}{$\ $pair} & \textcolor[HTML]{735173}{$\ $of} & \textcolor[HTML]{874963}{$\ $black} & \textcolor[HTML]{814C68}{$\ $scissors} & \textcolor[HTML]{944359}{$\ $and} & \textcolor[HTML]{9B4055}{$\ $a} & \textcolor[HTML]{A13D50}{$\ $pink} & \textcolor[HTML]{BC323B}{$\ $plate} \\
- & - & - & - & - & 2 & 2 & 13 & 12 & 15 & 14 & 17 & 18 & 19 & 23 \\\midrule
\end{tabular}}}

\makebox[\textwidth]{\resizebox{\textwidth}{!}{\begin{tabular}{cccccccccccccccccc}
\textcolor[HTML]{B53540}{$\ $with} & \textcolor[HTML]{C22F36}{$\ $a} & \textcolor[HTML]{C92C31}{$\ $pig} & \textcolor[HTML]{CF292C}{'s} & \textcolor[HTML]{D62728}{$\ $face} & \textcolor[HTML]{A83A4A}{$\ $on} & \textcolor[HTML]{AF3745}{$\ $it} & \textcolor[HTML]{3F689B}{.} & \textcolor[HTML]{456696}{$\ $
} & \textcolor[HTML]{1F77B4}{The} & \textcolor[HTML]{1F77B4}{$\ $common} & \textcolor[HTML]{1F77B4}{$\ $item} & \textcolor[HTML]{1F77B4}{$\ $in} & \textcolor[HTML]{1F77B4}{$\ $the} & \textcolor[HTML]{1F77B4}{$\ $two} & \textcolor[HTML]{1F77B4}{$\ $images} & \textcolor[HTML]{1F77B4}{$\ $} & \textcolor[HTML]{1F77B4}{$\ $is} \\
22 & 24 & 25 & 26 & 27 & 20 & 21 & 4 & 5 & - & - & - & - & - & - & - & - & - \\\midrule
\end{tabular}}}

\begin{tabular}{ccc}
\textcolor[HTML]{6D5478}{$\ $the} & \textcolor[HTML]{66577D}{$\ $scissors} & \textcolor[HTML]{605A82}{.} \\
11 & 10 & 9 \\\midrule
\end{tabular}
\end{minipage}
\end{tcolorbox}
\caption{Example: Structured Reasoning and Early Answering.}
\label{tab:exp1}
\vspace{-0.5em}
\end{table*}

\begin{table*}[t]
\centering
\begin{tcolorbox}[
  width=\linewidth,
  boxrule=3pt,
  colframe=lightgray,
  colback=lightgray!35,
  boxsep=1pt,
  left=3pt,
  right=3pt,
  top=1pt,
  bottom=1pt
]
\begin{minipage}{\textwidth}
\flushleft
\small
\begin{minipage}[b]{0.8\textwidth}
\textbf{Question}: Extract the information in the image.\\
\textbf{Response}: \{"Date : "Apr 18, 2018", "Time: "12:00 AM",\}\\
\textbf{Response Length/\#Remaining Tokens/Actual Iterations}: 32/22/7\\\\
\textbf{Generation History}:\\
\begin{tabular}{c@{$\ \ $}c@{$\ \ $}c@{$\ \ $}c@{$\ \ $}c@{$\ \ $}c@{$\ \ $}c@{$\ \ $}c@{$\ \ $}c@{$\ \ $}c@{$\ \ $}c@{$\ \ $}c@{$\ \ $}c@{$\ \ $}c@{$\ \ $}c@{$\ \ $}c@{$\ \ $}c@{$\ \ $}c@{$\ \ $}c@{$\ \ $}c@{$\ \ $}c@{$\ \ $}c@{$\ \ $}c@{$\ \ $}c@{$\ \ $}c@{$\ \ $}c@{$\ \ $}c@{$\ \ $}c@{$\ \ $}c@{$\ \ $}c@{$\ \ $}c}\midrule
\textcolor[HTML]{1F77B4}{\{
} & \textcolor[HTML]{1F77B4}{	} & \textcolor[HTML]{1F77B4}{"} & \textcolor[HTML]{1F77B4}{Date} & \textcolor[HTML]{1F77B4}{$\ \ $:} & \textcolor[HTML]{1F77B4}{$\ \ $"} & \textcolor[HTML]{A83A4A}{Apr} & \textcolor[HTML]{4C6290}{$\ \ $} & \textcolor[HTML]{4C6290}{1} & \textcolor[HTML]{4C6290}{8} & \textcolor[HTML]{4C6290}{,} & \textcolor[HTML]{4C6290}{$\ \ $} & \textcolor[HTML]{4C6290}{2} & \textcolor[HTML]{4C6290}{0} & \textcolor[HTML]{4C6290}{1} & \textcolor[HTML]{4C6290}{8} & \textcolor[HTML]{63587F}{",
} & \textcolor[HTML]{4C6290}{	} & \textcolor[HTML]{1F77B4}{"} & \textcolor[HTML]{1F77B4}{Time} & \textcolor[HTML]{1F77B4}{:} & \textcolor[HTML]{1F77B4}{$\ \ $"} & \textcolor[HTML]{7A4E6D}{1} & \textcolor[HTML]{7A4E6D}{2} & \textcolor[HTML]{91445C}{:} & \textcolor[HTML]{7A4E6D}{0} & \textcolor[HTML]{91445C}{0} & \textcolor[HTML]{BF3039}{$\ \ $AM} & \textcolor[HTML]{D62728}{",} & \textcolor[HTML]{1F77B4}{
} & \textcolor[HTML]{1F77B4}{\}} \\
- & - & - & - & - & - & 5 & 1 & 1 & 1 & 1 & 1 & 1 & 1 & 1 & 1 & 2 & 1 & - & - & - & - & 3 & 3 & 4 & 3 & 4 & 6 & 7 & - & - \\\midrule
\end{tabular}
\end{minipage}
\hspace{-0.5em}
\begin{minipage}[b]{0.2\textwidth}
\includegraphics[width=3cm, trim=0 20cm 0 22cm, clip]{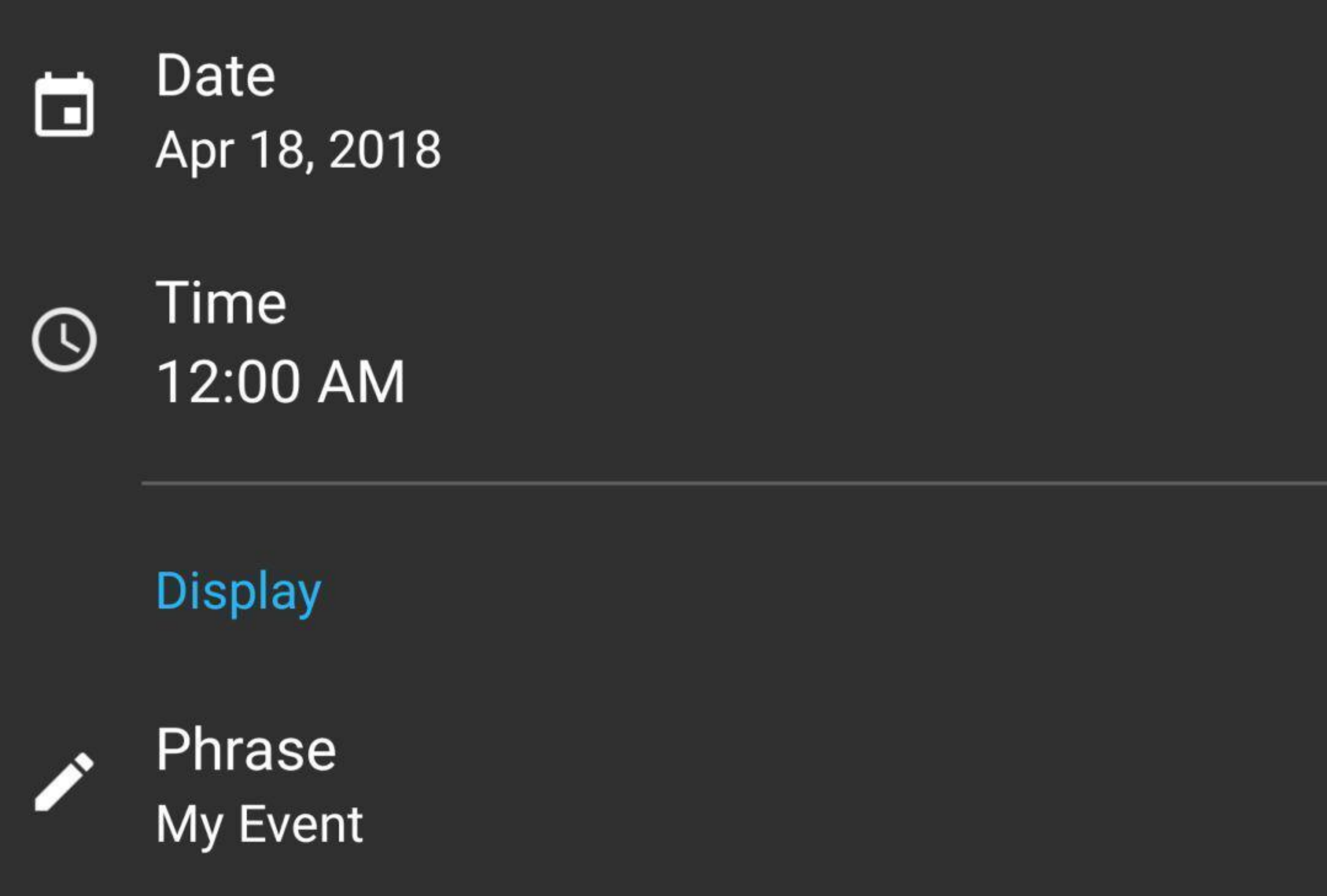}
\end{minipage}
\end{minipage}
\end{tcolorbox}
\caption{Example: Structured output and automatic semantic clustering during the generation process.
}
\label{tab:exp2}
\vspace{-0.5em}
\end{table*}

\begin{table*}[t]
\centering
\begin{tcolorbox}[
  width=\linewidth,
  boxrule=3pt,
  colframe=lightgray,
  colback=lightgray!35,
  boxsep=1pt,
  left=3pt,
  right=3pt,
  top=1pt,
  bottom=1pt
]
\begin{minipage}{\textwidth}
\flushleft
\small
\begin{minipage}[b]{0.63\textwidth}
\textbf{Question}: How many objects are preferred by more than 90 percent of people in at least one category? Please think step by step.

\vspace{0.5em}
\textbf{Response}: Nothing is preferred by more than 90 percent of people in any category.
Thus, the answer is \textbackslash box\{Zero\}.\\
\textbf{Response Length/\#Remaining Tokens/Actual Iterations}:\\
\centerline{32/23/16}\\
\textbf{Generation History}:
\end{minipage}
\hfill
\begin{minipage}[b]{0.35\textwidth}
\hspace{1.5em}
\includegraphics[height=2.7cm, trim=0 155 0 0, clip]{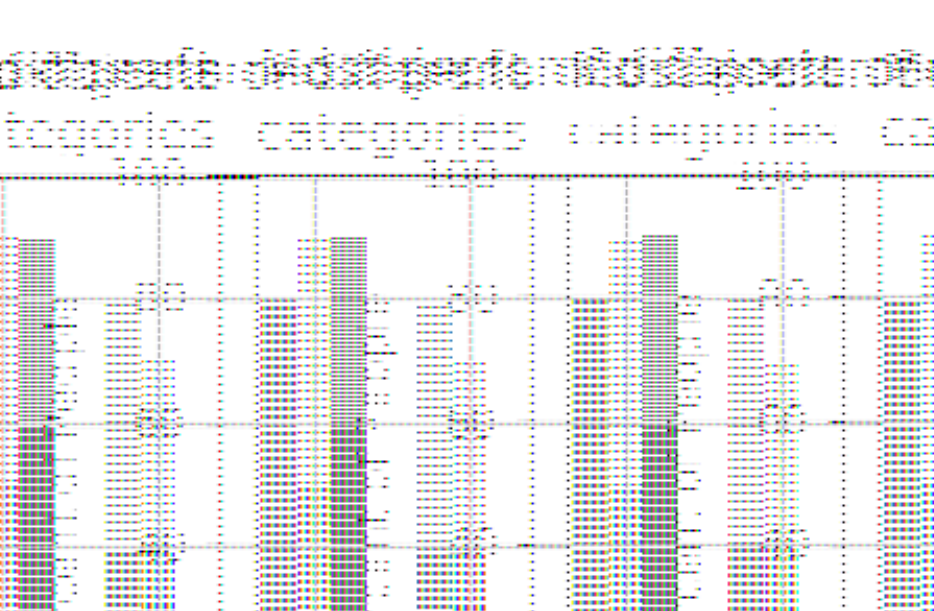}
\end{minipage}
\makebox[\textwidth]{\resizebox{\textwidth}{!}{\begin{tabular}{ccccccccccccccccc}\midrule
\textcolor[HTML]{D62728}{Nothing} & \textcolor[HTML]{CB2B2F}{$\ $is} & \textcolor[HTML]{C12F37}{$\ $preferred} & \textcolor[HTML]{B7343F}{$\ $by} & \textcolor[HTML]{AD3846}{$\ $more} & \textcolor[HTML]{A33D4E}{$\ $than} & \textcolor[HTML]{994156}{$\ $} & \textcolor[HTML]{8E465E}{9} & \textcolor[HTML]{8E465E}{0} & \textcolor[HTML]{8E465E}{$\ $percent} & \textcolor[HTML]{844A66}{$\ $of} & \textcolor[HTML]{7A4E6D}{$\ $people} & \textcolor[HTML]{705375}{$\ $in} & \textcolor[HTML]{66577D}{$\ $any} & \textcolor[HTML]{5C5C85}{$\ $category} & \textcolor[HTML]{476595}{.
} & \textcolor[HTML]{1F77B4}{Thus} \\
17 & 16 & 15 & 14 & 13 & 12 & 11 & 10 & 10 & 10 & 9 & 8 & 7 & 6 & 5 & 3 & - \\\midrule
\end{tabular}}}

\makebox[\textwidth][l]{\resizebox{0.4\textwidth}{!}{\begin{tabular}{ccccccccc}
\textcolor[HTML]{1F77B4}{,} & \textcolor[HTML]{1F77B4}{$\ $the} & \textcolor[HTML]{1F77B4}{$\ $answer} & \textcolor[HTML]{1F77B4}{$\ $is} & \textcolor[HTML]{1F77B4}{$\ $} & \textcolor[HTML]{1F77B4}{box} & \textcolor[HTML]{1F77B4}{\{} & \textcolor[HTML]{51608D}{Zero} & \textcolor[HTML]{3D699C}{\}.} \\
- & - & - & - & - & - & - & 4 & 2 \\\midrule
\end{tabular}}}

\vspace{0.5em}
\textbf{Response}: All of the objects in the chart have a maximum of 90 percent, but no object actually reaches 90. Therefore, none of the objects are preferred by more than 90 percent of people in at least one category. 
Thus, the answer is \textbackslash box\{Zero\}.\\
\textbf{Response Length/\#Remaining Tokens/Actual Iterations}: 64/55/37\\
\textbf{Generation History}:\\
\makebox[\textwidth]{\resizebox{\textwidth}{!}{\begin{tabular}{ccccccccccccccccc}\midrule
\textcolor[HTML]{814C68}{All} & \textcolor[HTML]{735173}{$\ $of} & \textcolor[HTML]{6A567A}{$\ $the} & \textcolor[HTML]{6E5477}{$\ $objects} & \textcolor[HTML]{77506F}{$\ $in} & \textcolor[HTML]{7D4D6C}{$\ $the} & \textcolor[HTML]{D62728}{$\ $chart} & \textcolor[HTML]{864964}{$\ $have} & \textcolor[HTML]{8A4761}{$\ $a} & \textcolor[HTML]{9E3F52}{$\ $maximum} & \textcolor[HTML]{8F455D}{$\ $of} & \textcolor[HTML]{994156}{$\ $} & \textcolor[HTML]{A73B4B}{9} & \textcolor[HTML]{944359}{0} & \textcolor[HTML]{A23D4F}{$\ $percent} & \textcolor[HTML]{AB3948}{,} & \textcolor[HTML]{BF3139}{$\ $but} \\
20 & 17 & 15 & 16 & 18 & 19 & 38 & 21 & 22 & 26 & 23 & 25 & 28 & 24 & 27 & 29 & 33 \\\midrule
\end{tabular}}}

\makebox[\textwidth]{\resizebox{\textwidth}{!}{\begin{tabular}{cccccccccccccccc}
\textcolor[HTML]{C32F36}{$\ $no} & \textcolor[HTML]{C82C32}{$\ $object} & \textcolor[HTML]{CC2B2F}{$\ $actually} & \textcolor[HTML]{D1282B}{$\ $reaches} & \textcolor[HTML]{B53540}{$\ $} & \textcolor[HTML]{BA333D}{9} & \textcolor[HTML]{B03744}{0} & \textcolor[HTML]{605A82}{.} & \textcolor[HTML]{65587E}{$\ $Therefore} & \textcolor[HTML]{5C5C85}{,} & \textcolor[HTML]{565E89}{$\ $none} & \textcolor[HTML]{565E89}{$\ $of} & \textcolor[HTML]{52608C}{$\ $the} & \textcolor[HTML]{496493}{$\ $objects} & \textcolor[HTML]{446697}{$\ $are} & \textcolor[HTML]{3F689B}{$\ $preferred} \\
34 & 35 & 36 & 37 & 31 & 32 & 30 & 13 & 14 & 12 & 11 & 11 & 10 & 8 & 7 & 6 \\\midrule
\end{tabular}}}

\makebox[\textwidth]{\resizebox{\textwidth}{!}{\begin{tabular}{ccccccccccccccccc}
\textcolor[HTML]{3F689B}{$\ $by} & \textcolor[HTML]{3F689B}{$\ $more} & \textcolor[HTML]{3F689B}{$\ $than} & \textcolor[HTML]{3F689B}{$\ $} & \textcolor[HTML]{3F689B}{9} & \textcolor[HTML]{3F689B}{0} & \textcolor[HTML]{3F689B}{$\ $percent} & \textcolor[HTML]{3F689B}{$\ $of} & \textcolor[HTML]{3A6A9E}{$\ $people} & \textcolor[HTML]{3A6A9E}{$\ $in} & \textcolor[HTML]{3A6A9E}{$\ $at} & \textcolor[HTML]{3A6A9E}{$\ $least} & \textcolor[HTML]{356CA2}{$\ $one} & \textcolor[HTML]{3A6A9E}{$\ $category} & \textcolor[HTML]{4D6290}{.} & \textcolor[HTML]{1F77B4}{Thus} & \textcolor[HTML]{1F77B4}{,} \\
6 & 6 & 6 & 6 & 6 & 6 & 6 & 6 & 5 & 5 & 5 & 5 & 4 & 5 & 9 & - & - \\\midrule
\end{tabular}}}

\makebox[\textwidth][l]{\resizebox{0.4\textwidth}{!}{\begin{tabular}{cccccccc}
\textcolor[HTML]{1F77B4}{$\ $the} & \textcolor[HTML]{1F77B4}{$\ $answer} & \textcolor[HTML]{1F77B4}{$\ $is} & \textcolor[HTML]{1F77B4}{$\ $} & \textcolor[HTML]{1F77B4}{box} & \textcolor[HTML]{1F77B4}{\{} & \textcolor[HTML]{316EA5}{Zero} & \textcolor[HTML]{2C71A9}{\}.} \\
- & - & - & - & - & - & 3 & 2 \\\midrule
\end{tabular}}}
\end{minipage}
\end{tcolorbox}
\vspace{-0.3em}
\caption{Example: Length Control.}
\label{tab:exp3}
\vspace{-1em}
\end{table*}

In this section, we provide a showcase to demonstrate that dMLLMs exhibit fundamentally different decoding behaviors compared to autoregressive models. 
Unlike autoregressive models, discrete diffusion models offer the unique capability of \textbf{explicitly controlling the content} generated at arbitrary positions. This allows us to specify certain tokens as fixed in the final answer prior to generation. We refer to such tokens as \textbf{structure priors}.

We present three illustrative examples. Each includes the input image, the question, and the generated response from Dimple. For each case, we annotate: the \textbf{response length} (i.e., the total number of tokens in the final answer), the \textbf{number of remaining tokens} that need to be generated after applying the structure prior, and the \textbf{actual number of decoding steps} (\textit{actual iteration}) used in generation.
Furthermore, we visualize the \textbf{generation history}, which records at which iteration each token was generated. Tokens provided by structure priors are marked with ``-'', and special tokens (e.g., padding) are excluded from the generation history but counted in the actual number of decoding steps. To enhance interpretability, we colored the response: tokens decoded earlier are shown in blue, while those generated later appear in red. Tokens with similar decoding steps share similar colors. All examples leverage the Prefilling technique and proposed Confident Decoding strategy.

\subsection{Example 1 (\cref{tab:exp1}): Structured Reasoning and Early Answering}

In this example, the input consists of two images, and the model is asked to identify the common object present in both. We inject three structure priors into the expected answer: ``In the first image, there '', ``In the second image, there '' and ``The common item in the two images is''.
These priors guide the model to follow a reasoning trajectory resembling structured thought. Unlike the chain-of-thought (CoT) strategies used in autoregressive models, which rely on indirect guidance, structure prior enables direct and precise control over intermediate reasoning steps.

Remarkably, the correct answer ``scissors'' is decoded at the \textbf{10\textsuperscript{th}} iteration, prior to the completion of the full response. The subsequent steps focus on completing the thinking trajectory. This illustrates a key strength of diffusion models: they can arrive at the final answer earlier in the generation process. In contrast, autoregressive models must complete the full sequence before producing the answer.

\subsection{Example 2 (\cref{tab:exp2}): Structured Output}

In this example, the task involves extracting visual information from an image. The question does not specify the desired output format and the attributes required. All of these are indicated by the structure prior. In this example, the structure prior is ``\{date:'', ``time:'' and ''\}''.
These priors define a JSON-like response layout, and the model successfully generates structured content accordingly. 

This example also provides direct evidence of the effectiveness of \textbf{Confident Decoding}. The number of tokens decoded per iteration varies dynamically. For instance, 9 tokens corresponding to the date and year are decoded simultaneously in a single step. Despite a total of 22 tokens to be generated, the model completes generation in only 7 iterations, one-third of the total number of tokens. This parallelism significantly accelerates inference.

\subsection{Example 3 (\cref{tab:exp3}): Length Control}
Controlling output length is difficult in autoregressive generation. However, discrete diffusion models inherently support this capability. In this case, we specify structure priors at the end of the sequence. Specifically, we force tokens at positions \( [-12:-4] \) to be:
``Thus, the answer is \textbackslash box\{''.
This allows us to strictly control where the model concludes its generation and places the final answer. We demonstrate two configurations with response lengths of 16 and 32. In each case, the model adjusts its reasoning span to fill the available token budget appropriately. 

\section{Related Work}

In the supplementary material, we also discuss the differences between Dimple and other approaches based on parallel decoding and multimodal perception models.

The first problem when introducing diffusion process into language decoding, is the choice of a continuous diffusion space or discrete one. There are explorations in the continuous spaces.
DiffuSeq \citep{diffuseq} use a partial noising and conditional denoising process to model sequence-to-sequence diffusion in continuous space. Argmax Flows and Multinomial Diffusion propose continuous relaxations for categorical distributions for better generation stability \citep{argmax-flows}. SED \citep{self-conditioned-embedding} directly use the embedding space for the diffusion of natural language tokens.  
For the exploration in the discrete space,
Structured Denoising Diffusion Models (SDDM) introduce discrete diffusion tailored to linguistic structures \citep{sddm}; RDM \citep{reparam-discrete-diffusion} reparameterizes the backward discrete diffusion into a step sampling process to provide more flexible and unified framework. 
For training diffusion language model, loss function is the key. MDLM \citep{simple-masked-diffusion}, MD4~\citep{md4} and SEDD \citep{discrete-diffusion} show that both a simple weighted mask language modeling loss and a score entropy loss, which is adopted from the score matching loss, are suitable for training.
In this work, we follow the thread of masked language modeling scheme for the language diffusion process. In this direction,
Diffusion-LM \citep{diffusion-lm} applies diffusion to masked language modeling to enhance controllable text generation; DiffusionBERT \citep{diffusion-bert} uses pretrained BERT as an initialization of the diffusion training to accelerate convergence and improve performance. 
Besides using a pure diffusion language model,
auto-regressive hybrids like AR-Diffusion \citep{ar-diffusion} and Semi-autoregressive methods like SSD-LM \citep{ssd-lm} aim to combine the benefits of diffusion and autoregressive modeling. SSD-LM also shows that with a simplex projection the diffusion process can also be conducted in the natural vocabulary space.
Recently, diffusion language models have been scaled to show comparable performance with the SOTA auto-regressive model, to learn instruction following ability and to have even better planning ability~\citep{llada,diffullama,ye2025autoregressiondiscretediffusioncomplex,dream}.

\vspace{-0.5em}
\section{Conclusion}
\vspace{-0.5em}
In this work, we introduce Discrete Diffusion Multimodal Large Language Models \textbf{Dimple} and \textbf{Dimple+}, both of which match the performance of their autoregressive baselines while offering unique advantages such as parallel decoding and structure-aware generation. 
The main limitation of Dimple is that Dimple-7B is trained on less than 1/20th of the data used by state-of-the-art models such as Qwen2.5-VL and Eagle2. Consequently, while Dimple demonstrates competitive performance among models of similar scale, it still falls short compared to the most powerful models.

\section*{Acknowledgment}
This project is supported by the Ministry of Education, Singapore, under the Academic Research Fund Tier~1 (FY2026, WBS: A-8004345-00-00),  the National Research Foundation, Singapore, and Cyber Security Agency of Singapore under its National Cybersecurity R\&D Programme and CyberSG R\&D Cyber Research Programme Office (Award: CRPO-GC1-NTU-002).

\bibliography{main}

\clearpage

\onecolumn
\renewcommand{\thefigure}{S\arabic{figure}}
\renewcommand{\thetable}{S\arabic{table}}
\renewcommand{\thesection}{S\arabic{section}}
\setcounter{figure}{0}
\setcounter{table}{0}
\setcounter{section}{0}
{
    \centering
    \Large
    \vspace{0.5em}
    \textbf{\textit{- Supplementary Material -}}\\
    \vspace{0.1em}
}

\section{Training and Evaluation}
\subsection{Training Configuration and Hyperparameters}

\begin{table}[h]
\centering 
\small 
\renewcommand{\arraystretch}{1.3} 
\makebox[\textwidth]{\resizebox{1.0\textwidth}{!}{
\begin{tabular}{@{}lccc@{}}
\toprule
Training Phase          & \multicolumn{1}{l}{1st: Autoregressive Alignment} & \multicolumn{1}{l}{2nd: Autoregressive Instruction Tuning} & \multicolumn{1}{l}{3rd: Diffusion Instruction Tuning} \\ \midrule
Training Method         & Autoregressive                                    & Autoregressive                                             & Diffusion                                             \\
Training Data           & LLaVA-CC3M-Pretrain                               & \multicolumn{2}{c}{LLaVA-NEXT Instruction Tuning}                                                                  \\
\#Training Data & 559k                                              & \multicolumn{2}{c}{739k}                                                                                           \\
learning rate           & 0.001                                            & 5e-6                                                      & 5e-7                                                 \\
batch size              & 256                                               & 128                                                        & 128                                                   \\
weight decay            & 0                                                 & 0                                                          & 0                                                     \\
warmup ratio            & 0.03                                              & 0.03                                                       & 0.03                                                  \\
lr schedule             & linear                                            & linear                                                     & linear                                                \\
max grad norm           & 1                                                 & 1                                                          & 0.1                                                   \\
epoch                   & 1                                                 & 1                                                          & 1                                                     \\
optimizer               & AdamW                                             & AdamW                                                      & AdamW                                                 \\ \bottomrule
\end{tabular}}}
\vspace{1em}
\caption{Training Configuration and Hyper-Parameters.
}
\label{tab:training_hyper}
\end{table}
The training of Dimple can be divided into three stages. The first two stages involve autoregressive alignment and instruction-following tuning, while the final stage utilizes diffusion-based instruction-following tuning. The specific training configurations and hyperparameters for each stage are detailed in \cref{tab:training_hyper}. 
The training hyperparameters of Dimple+ are the same as those used in the DT stage of Dimple.
The training of Dimple and Dimple+ was conducted on H100 clusters, requiring approximately 100 GPU hours each.
Experiments in \cref{tab:prefill} is conducted on one single H100 GPU. 

The number of training iterations for the Dimple AR baseline equals the sum of the iterations used in the autoregressive training and diffusion training stages of Dimple. The AR baseline for Dimple+ is trained with the same number of iterations as Dimple+. For both cases, the training hyperparameters of the AR baselines are aligned with those used in the autoregressive training stage of Dimple.

\subsection{Masking Strategy}
During training, the Autoregressive Alignment and Autoregressive Tuning phases use the standard next-token prediction loss from AR training. In these stages, input samples are not masked.  In contrast, during the Diffusion Tuning stage, the response parts of the input samples are masked. The masking strategy follows a linear schedule, where the proportion of masked tokens is proportional to the diffusion time step. Let $r \in [0,1]$ and $t \in [0,1]$ denote the masking ratio and the time index, respectively. We define $r = t$. That is, when $t = 0$, no tokens are masked, and when $t = 1$, all tokens are masked. This linear masking strategy is also commonly used in prior discrete diffusion LLM training ~\citep{dream,gong2024scaling,d3pm}.

\subsection{Training Sample Padding}
\begin{table}[h]
\centering 
\small 
\renewcommand{\arraystretch}{1.3} 
\makebox[\textwidth]{\resizebox{1.0\textwidth}{!}{
\begin{tabular}{ll}
\hline
\multicolumn{2}{l}{Text Input for Autoregressive Training}                                                                                                                           \\ \hline
User:      & \texttt{[BOS]}\textless{}image\textgreater{}\textbackslash{}nWhat type of item is holding the napkin?\textbackslash{}nAnswer the   question with GPT-T-COCO format.\texttt{[EOS]} \\
Assistant: & \texttt{[BOS]}A clear napkin holder is holding the napkin.\texttt{[EOS]}                                                                                                          \\
User:      & \texttt{[BOS]}What color is the napkin in the napkin holder?\texttt{[EOS]}                                                                                                        \\
Assistant: & \texttt{[BOS]}The napkin inside the napkin holder is blue-green.\texttt{[EOS]}                                                                                                    \\ \hline
\multicolumn{2}{l}{Text Input   for Diffusion Training}                                                                                                                              \\ \hline
User:      & \texttt{[BOS]}\textless{}image\textgreater{}\textbackslash{}nWhat type of item is holding the napkin?\textbackslash{}nAnswer the   question with GPT-T-COCO format.\texttt{[EOS]} \\
Assistant: & \texttt{[BOS]}A clear napkin holder is holding the napkin.$\underbrace{\texttt{[Padding]}\dots\texttt{[Padding]}}_{n_1\text{ \texttt{[Padding]} Tokens}}$                                                                                      \\
User:      & \texttt{[BOS]}What color is the napkin in the napkin holder?\texttt{[EOS]}                                                                                                        \\
Assistant: & \texttt{[BOS]}The napkin inside the napkin holder is   blue-green.$\underbrace{\texttt{[Padding]}\dots\texttt{[Padding]}}_{n_2\text{ \texttt{[Padding]} Tokens}}$                                                                           \\ \hline
\end{tabular}}}
\vspace{1em}
\caption{Training Examples Used for Autoregressive Training and Diffusion Training. $n_1$ and $n_2$ n 
are random integers that enhance the model's robustness to varying response lengths.
}
\label{tab:training_example}
\end{table}

Unlike autoregressive models, dMLLMs generate answers with a given response length and do not rely on a \texttt{[EOS]} token to terminate the generation. 
To accommodate the diffusion generation process, we replace the \texttt{[EOS]} token in the original training dataset with a random number of padding tokens specific to Dream's tokenizer. These tokens are used by the Dream model to learn to adaptively control the actual length of the generated answers. 

The Dream model relies on a predefined response length to determine the length of generated outputs, ensuring that the number of textual tokens does not exceed this length. When the number of textual tokens is less than the response length, Dream generates additional \texttt{[Padding]} tokens so that the total number of tokens equals the predefined response length. Therefore, the model does not rely on a \texttt{[EOS]} token to terminate generation. Instead, \texttt{[EOS]} tokens in the original training data should be replaced with \texttt{[Padding]} tokens.

To enhance the Dimple model's robustness to response length, we randomly replace each single \texttt{[EOS]} token with $n$ \texttt{[Padding]} tokens. The max and min value of $n$ is selected based on the length of the ground truth response. 
For a ground truth answer with the length of $l$, the derivation of $n$ can be written as follows.
\begin{align*}
    &n\sim \text{RandomInteger}(n_{min}, n_{max}),\\
    & (n_{min}, n_{max}) = \begin{cases}
        (l + 1,\ 2^{\lceil \log_2(l + 1) \rceil}), & \text{if } n < 16, \\
        (l + 1,\ 16 \cdot \left\lceil \frac{l + 1}{16} \right\rceil), & \text{otherwise}
    \end{cases}
\end{align*}
\cref{tab:training_example} show an example of input sample for diffusion training. 

\begin{table}[t]
\centering

\begin{tcolorbox}[colback=gray!5!white, colframe=gray!40, title=\textbf{\hspace{-1em}Caption Evaluation Instructions}, coltitle=black]
You will be given one image and one generated visual caption describing the image.

Your task is to carefully evaluate how well the caption reflects the content of the image based on four dimensions:

\begin{enumerate}
    \item Accuracy: Does the caption correctly describe the objects, actions, or scene in the image?
    \item Completeness: Does the caption capture the essential and important aspects of the image, without omitting critical content?
    \item Relevance: Is all the information in the caption relevant to the image? Are there any hallucinated, fabricated, or irrelevant elements?
    \item Conciseness: Is the caption free from redundancy or unnecessary elaboration?
\end{enumerate}

For each dimension, assign a score from 0 to 100 and present it in the following format:

\vspace{0.5em}
```\\
**Accuracy**  \\
Justification: \{\{your explanation here\}\}\\
Accuracy score: \{\{score\}\}

\vspace{0.5em}
**Completeness**  \\
Justification: \{\{your explanation here\}\}\\
Completeness score: \{\{score\}\}

\vspace{0.5em}
**Relevance**  \\
Justification: \{\{your explanation here\}\}\\
Relevance score: \{\{score\}\}

\vspace{0.5em}
**Conciseness**  \\
Justification: \{\{your explanation here\}\}\\
Conciseness score: \{\{score\}\}\\
'''

\vspace{0.5em}
Then, provide the average score over the four dimensions. Justify your overall score in a paragraph and finish with the sentence: \\
```The final score is \{\{score\}\}.'''

\vspace{0.5em}
Image: [image]\\
Generated caption: [caption]
\end{tcolorbox}

\caption{Instructions used when Evaluating Captions for the Models after Alignment.}
\label{tab:cap_eval}
\end{table}
\subsection{Evaluation Configuration}
We conducted benchmark evaluations using the lmms-eval library. Both the evaluation metrics and prompts were based on the default configuration provided by lmms-eval. Throughout all evaluations, we employed the MaskGIT decoding algorithm from Dream, with the algorithm temperature set to 0 and confident decoding disabled. The overall decoding temperature was set to 0, and the top-p value was fixed at 0.01, resulting in a largely deterministic generation process.

The most influential parameter affecting evaluation performance is the response length. For all datasets except MMVet, we experimented with response lengths of 4, 8, and 16. Table 1 reports the best results among these three settings. The MMVet dataset is an exception due to the high variability in the ground truth response lengths—ranging from a single word to over 500 tokens. For MMVet, we fixed the token length for all responses to 64. While this is not an optimal choice, as it limits the model’s capacity to fully address long-response questions and wastes tokens on short responses, it provides a uniform evaluation setting. We consider adaptive response length selection as a direction for future research.
During evaluation, the number of generation steps was set equal to the response length, enforcing the model to decode exactly one token per step.

Following the default configuration, the evaluation metrics used for each benchmark are listed as follows.
\begin{itemize}
    \item \textbf{Accuracy}: GQA, MMBench, POPE, MMMU, SQA, AI2D, ChartQA, TextQA
    \item \textbf{Number of Correct Answer}: MME, OCRBench
    \item \textbf{GPT-Based Accuracy}: MathVista, MMVet
\end{itemize}

We collect the number in the Table 1 as follows. Dimple's, Dimple+'s and their AR Baselines' performance is obtained using lmms-eval library, which is also used in LLaVA~\citep{lmms_eval2024,zhang2024lmmsevalrealitycheckevaluation}. 
For Qwen2.5-VL, results are collected from their official reports and VLMEvalKit~\citep{duan2024vlmevalkit}. 
For LLaVA-NEXT, results on OCRBench are taken from \citep{eagle}; results on MathVista test-mini, ChartQA, and AI2D are obtained using lmms-eval.
Other results for LLaVA-NEXT and the results for other models are collected from their official reports. 
The training data information is collected from the respective official reports.

\subsection{Evaluation of Model after Alignment}

To evaluate the captioning capabilities of the model after alignment, we employed GPT-4o for assessment. The prompt used for evaluation is shown in \cref{tab:cap_eval}.

\subsection{Model and Data Used}
The Qwen2.5 and Qwen2.5-VL series is released with explicit documentation regarding model scale, training data composition, language coverage, and licensing conditions. The model family provides broad multilingual support, including Chinese and English. Most open-weight Qwen2.5 and Qwen2.5-VL models are released under the Apache 2.0 License.

The LLaVA-CC3M-Pretrain dataset is a visual instruction pretraining dataset derived from CC3M and related captioning resources. Its dataset card identifies the modality as image-text data and the language as English. The dataset documentation states that the data is intended to support research on large multimodal models and to facilitate reproduction of the LLaVA work. It also specifies that users must comply with the original licenses and terms of the underlying CC3M and BLIP resources. CC3M permits use of the dataset while requesting acknowledgement of Google LLC as the data source and providing the data on an “as is” basis. In this paper, our use of LLaVA-CC3M-Pretrain Data is solely for academic research and is consistent with the stated purpose and licensing requirements of the dataset.

The LLaVA-NeXT instruction tuning data is associated with the LLaVA-NeXT project, which emphasizes that its datasets and checkpoints are subject to the licenses and terms of their original sources. These include, where applicable, the OpenAI Terms of Use, LLaMA community or research licenses, Tongyi Qianwen research license terms, and the licenses of underlying vision-language datasets. The LLaVA-NeXT documentation further describes related evaluation and recaptioned data, including English and Chinese image captioning examples and video captioning data derived from ActivityNet, while noting that some evaluation data is intended for internal development rather than formal benchmarking. In this work, our use of LLaVA-NeXT instruction tuning resources follows the relevant original license conditions and remains within the academic research context for which these resources were released.

\section{Confident Decoding}
\begin{algorithm}[t]
\caption{Confident Decoding}
\label{alg:confident_decoding}
\tcp{Take temperature as an example of probability adjustment.}
\tcp{Confidence is calculated in a MaskGIT way.}
\tcp{Use random selection callback.}
\KwIn{Input token sequence \( \mathbf{x}_{t} \), logits \( \mathbf{z}_t \), temperature \( \tau \),  threshold \( \gamma \), number of positions \( K \) to select if the fallback is triggered, number of masked tokens \( N \)}
\KwOut{Updated token sequence \( \mathbf{x}_{t+1} \)}
\BlankLine
\tcp{Step 1: Sample token predictions and compute confidence}
\tcp{Pre-revision probabilities}
\( \mathbf{p}_t \leftarrow \text{softmax}(\mathbf{z}_t) \) \;
\tcp{Post-revision probabilities}
\( \tilde{\mathbf{p}}_t \leftarrow \text{softmax}(\mathbf{z}_t / \tau) \) \;
\ForEach{\( i\in \{1, \dots, N \}\)}{
    \tcp{MaskGIT confidence}
    \( c^{(i)} = \max(\mathbf{p}_{t}^{(i)}) \) \;
    \tcp{sampling}
    \( \tilde{x}_{t}^{(i)} \sim \text{Categorical}(\tilde{\mathbf{p}}_{t}^{(i)}) \) \;
}

\tcp{Step 2: Position selection}
\BlankLine
\( x_{t+1} \leftarrow x_t \) \;
\uIf{\( \exists i, \; c^{(i)} \geq \gamma \)}{
    \tcp{Step 2.1: Confident decoding}
    \( \mathcal{I} \leftarrow \{i \mid c^{(i)} \geq \gamma, i \in \{1, \dots, N\} \} \) \;
    \ForEach{\( i \in \mathcal{I} \)}{
        \( x_{t+1}^{(i)} \leftarrow \tilde{x}_t^{(i)} \) \;
    }
}
\Else{
    \tcp{Step 2.2: Fallback: decoding via random selection}
    Randomly select \( K \) positions: \( \mathcal{I} \leftarrow \text{RandomSample}(\{1, \dots, N\}, K) \) \;
    \ForEach{\( i \in \mathcal{I} \)}{
        \( x_{t+1}^{(i)} \leftarrow \tilde{x}_t^{(i)} \) \;
    }
}
\Return \( x_{t+1} \)
\end{algorithm}

We propose \textbf{Confident Decoding}, a token selection strategy designed for diffusion-based generation. Unlike traditional decoding methods that fix the number of tokens to be updated per iteration, Confident Decoding dynamically selects positions to decode based on the model's confidence in its predictions.

\paragraph{Overview.} At each decoding step, the model evaluates the confidence of its predictions across all masked positions. If any of the confidence scores exceed a predefined threshold \( \gamma \), the corresponding tokens are decoded all together. Otherwise, a fallback strategy is used to ensure generation continues in accordance with the total decoding budget. 
That is, if no confidence score exceeds the threshold, the second stage of selection is triggered, and another decoding algorithm can be invoked. 
Thus, the complete decoding pipeline can combine Confident Decoding with MaskGIT, Random selection, or other strategies. Alg.~\ref{alg:confident_decoding} presents an example of the pseudocode for Confident Decoding.
This mechanism enables adaptive decoding, leading to both improved generation quality and computational efficiency.

\paragraph{Confidence Estimation.} Confident Decoding does not impose any constraints on the method for computing confidence. 
It can use the maximum prediction probability as the confidence value,  as used in  MaskGIT, the entropy of the prediction distribution, or the margin between the highest and second-highest prediction probabilities. In general, the confidence can be expressed as a function 
$c=\phi(p)$, where $p$ is a prediction probability vector of length $V$, and the output is a scalar confidence score.

Since the prediction probabilities in LLMs can be revised by decoding parameters such as temperature, top-p and top-k, prior work has considered two options for choosing the input probability $p$: either before~\citep{llada} or after such revisions~\citep{dream}. In our experiments, we use the prediction probabilities before these revisions to compute confidence. This choice is motivated by the fact that adjustments, like temperature and top-p, are applied in a position-wise manner, without accounting for inter-positional relationships or preserving the relative confidence across different positions.
For example, in high-temperature or low top-p settings, the revised probability may become almost deterministic, where one position has a probability of approximately 1 and all others are 0. Such a probability is uninformative for computing confidence, as all positions would yield identical confidence scores.

\begin{figure}[t]
    \centering
    \makebox[\textwidth]{\resizebox{1\textwidth}{!}{
    \begin{minipage}[t]{\textwidth}
        \centering
        \begin{subfigure}[t]{0.46\textwidth}
            \centering
            \includegraphics[height=3.5cm]{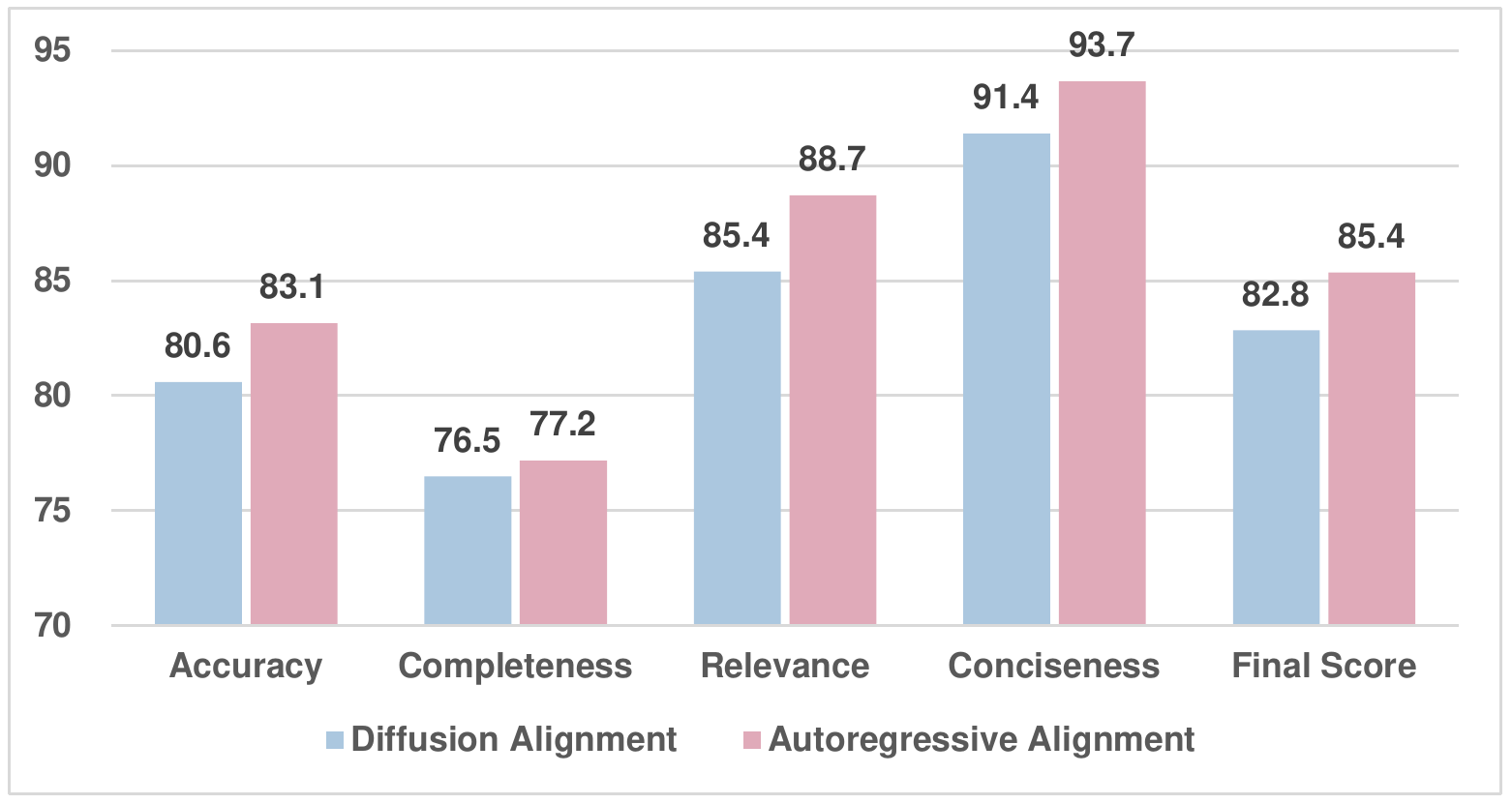}
            \caption{Performance after Alignment.}
            \label{fig:auto-abl:1}
        \end{subfigure}%
        \hfill
        \begin{subfigure}[t]{0.25\textwidth}
            \centering
            \includegraphics[height=3.5cm]{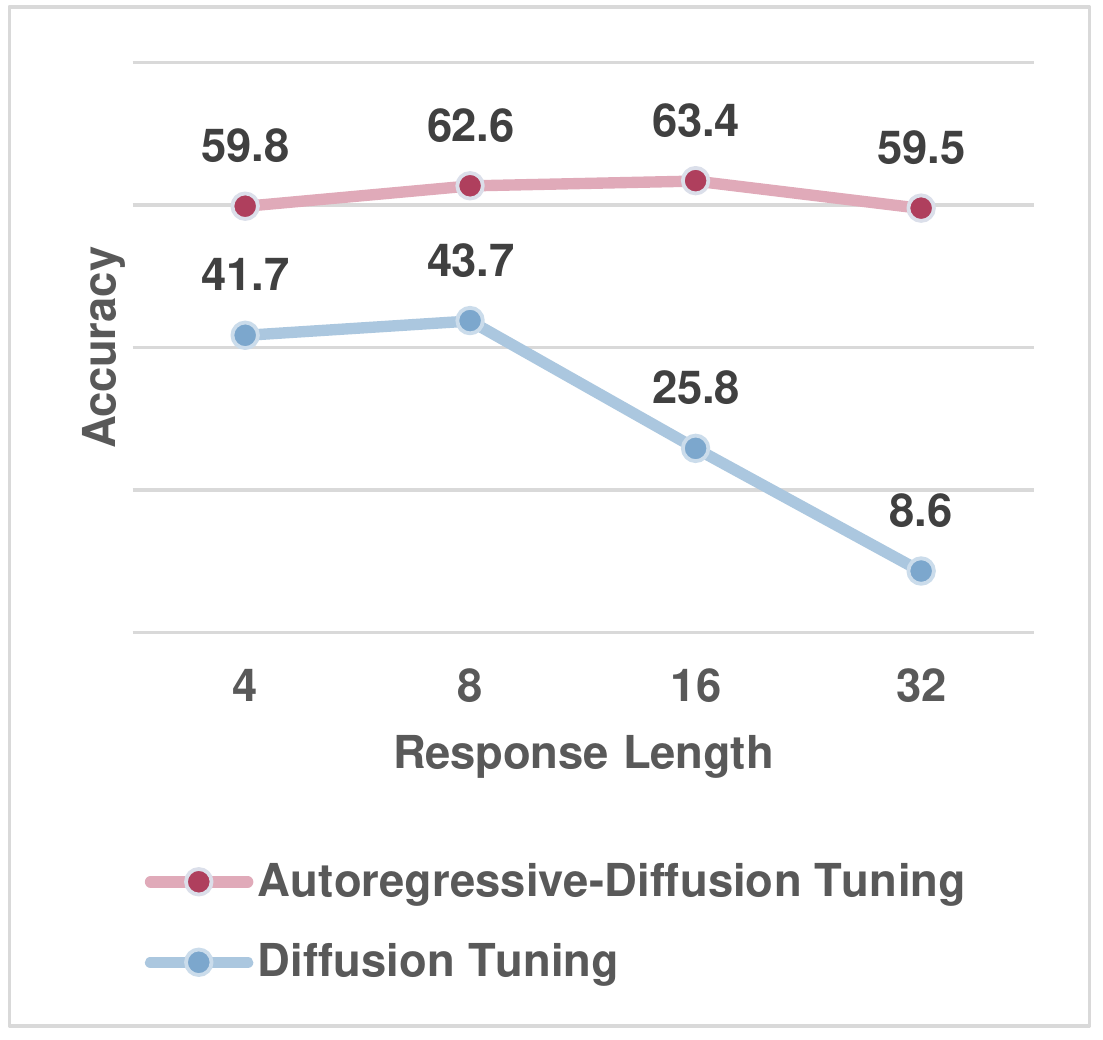}
            \caption{Length Bias: ChartQA.}
            \label{fig:auto-abl:2}
        \end{subfigure}%
        \hfill
        \begin{subfigure}[t]{0.25\textwidth}
            \centering
            \includegraphics[height=3.5cm]{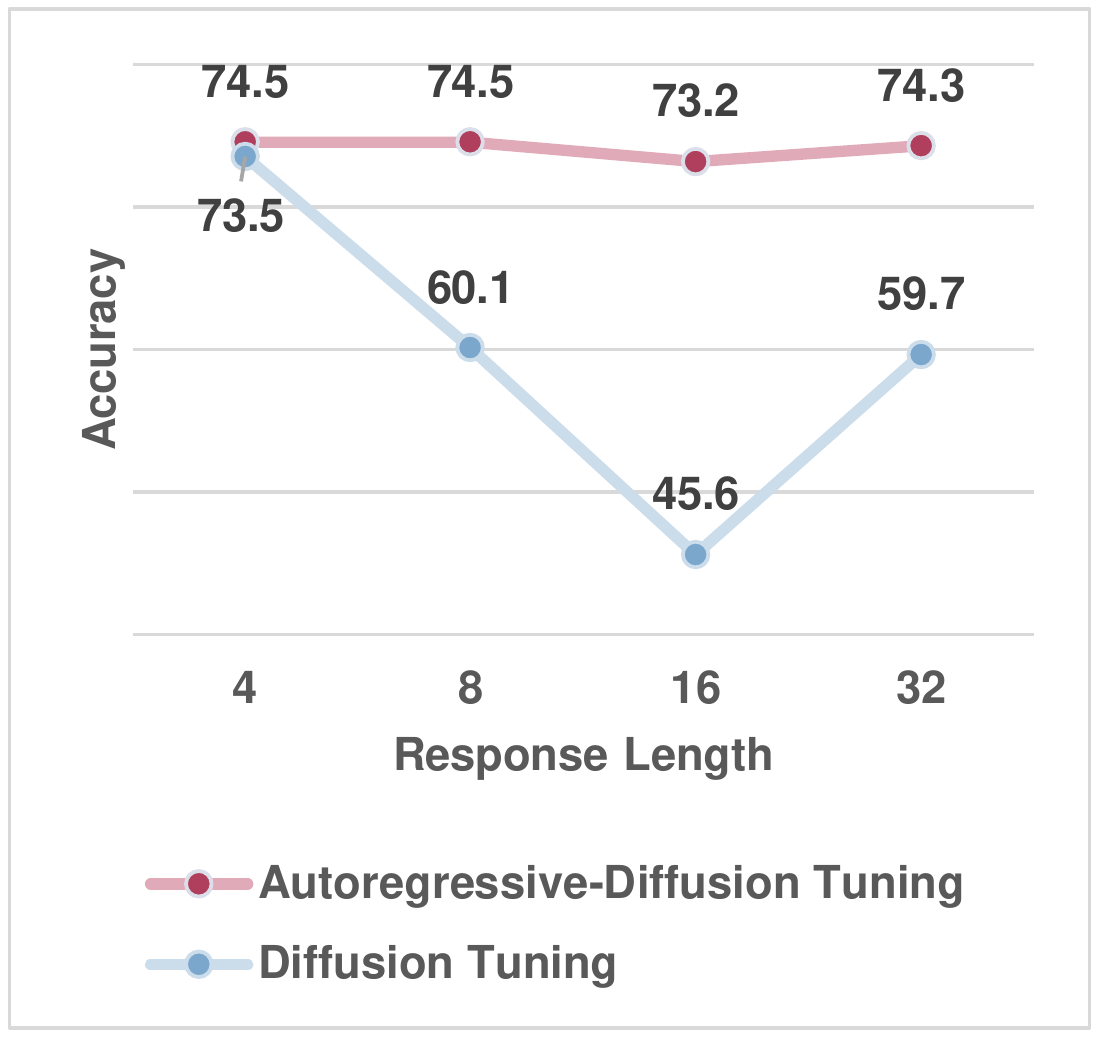}
            \caption{Length Bias: AI2D.}
            \label{fig:auto-abl:3}
        \end{subfigure}
        \caption{Comparison of Captioning Capabilities After Alignment (\cref{fig:auto-abl:1}), and the Length Bias Phenomenon on the ChartQA and AI2D Datasets After Instruction Tuning (\cref{fig:auto-abl:2,fig:auto-abl:3}).}
        \label{fig:auto-abl}
    \end{minipage}}}
\vspace{-1em}
\end{figure}
\section{Ablation on Dimple Training}
Following experiments demonstrates the advantages of the proposed autoregressive-then-diffusion hybrid training paradigm over the pure diffusion training. The pure diffusion pipeline consists of 2 phases: Diffusion Alignment (DA) and Diffusion Tuning (DT). Our hybrid approach incorporates 3 phases: Autoregressive Alignment (AA), Autoregressive Tuning (AT), and Diffusion Tuning (DT).

\paragraph{After Alignment}
To evaluate the captioning capability of the aligned models, we randomly sampled 128 images from the LLaVA-CC3M dataset~\citep{llava} (excluded from training) and generated captions using the aligned models. The model trained with DA used the diffusion generation pipeline, while the model trained with AA used the autoregressive generation pipeline. Given an image and its caption, we prompted GPT-4o-2024-11-20 to rate the captions on four dimensions—accuracy, completeness, relevance, and conciseness—on a scale of 1–100, with the final score being the average of the four dimensions. The results, summarized in \cref{fig:auto-abl:1}, show that the model aligned using AA outperforms the one aligned using DA across all metrics.

\begin{table}[t]
\centering 
\small 
\renewcommand{\arraystretch}{1.3} 
\makebox[\textwidth]{\resizebox{1\textwidth}{!}{
\begin{tabular}{@{}ll@{$\ $}ccccccccc@{}}
\toprule
                                          && TextVQA & SQA  & MMMU & \multicolumn{2}{c}{MME} & POPE & MMVet & ChartQA & AI2D \\ \midrule
A&DT (lr=5e-6, 1   Epoch)     & 38.4    & 70.2 & 45.1 & 1420            & 360            & 83.9 & 35.2  & 40.8    & 71.3 \\
B&DT (lr=5e-6, 2   Epoch)     & 36.6    & 70.6 & 44.0 & 1508            & 397            & 84.7 & 34.9  & 43.7    & 73.5 \\ \midrule
C&AT (lr=2e-5, 1 Epoch)  & 49.2    & 79.5 & 38.1 & 1298            & 403            & 84.3 & 32.9  & 53.7    & 70.5 \\
E&$\quad$+ DT (lr=5e-6, 1 Epoch) & \textbf{49.8}    & 81.9 & 42.6 & 1483            & 426            & \textbf{85.3} & 34.1  & 53.9    & \textbf{74.4} \\ \midrule
D&AT (lr=5e-6, 1 Epoch)  & \textbf{49.8}    & 83.7 & 43.3 & 1404            & 422            & 82.6 & 36.3  & 55.6    & 71.8 \\
F&$\quad$+ DT (lr=5e-7, 1 Epoch) & 49.4    & \textbf{84.4} & \textbf{45.2} & \textbf{1514 }           & \textbf{432}            & 85.1 & \textbf{41.2}  & \textbf{58.4}    & \textbf{74.4} \\ \bottomrule
\end{tabular}}}
\vspace{0.5em}
\caption{Performance Comparison Between Autoregressive Tuning (AT) and Diffusion Tuning (DT). 
The results show that the best performance is achieved by first applying autoregressive tuning followed by diffusion tuning.
}
\vspace{-2em}
\label{tab:auto-diff}
\end{table}
\paragraph{After Instruction Tuning}
We compared six different instruction tuning strategies:
\begin{itemize}[leftmargin=20pt, itemsep=0pt, parsep=0pt]
    \item Strategies A, B: Continue with DT for 1 or 2 epochs after DA.
    \item Strategies C, D: Perform 1 epoch of AT with two different learning rates after AA.
    \item Strategies E, F: Continue with 1 epoch of DT using two different learning rates after C and D.
\end{itemize}

\cref{tab:auto-diff} presents the performance of all six strategies. Since the final goal is to obtain a model capable of diffusion generation, all models were evaluated using the diffusion generation pipeline. First, comparing strategies A and D, which are both trained with the same learning rate for 1 epoch, AT outperforms DT on 7 out of 9 benchmarks. However, models trained with AT alone suffer from a gap between autoregressive training and diffusion-based inference. Then, comparing strategies B and F, the AT-then-DT strategy outperform the pure diffusion tuning strategy across all benchmarks.

\begin{figure}
    \centering
    \includegraphics[width=0.6\textwidth]{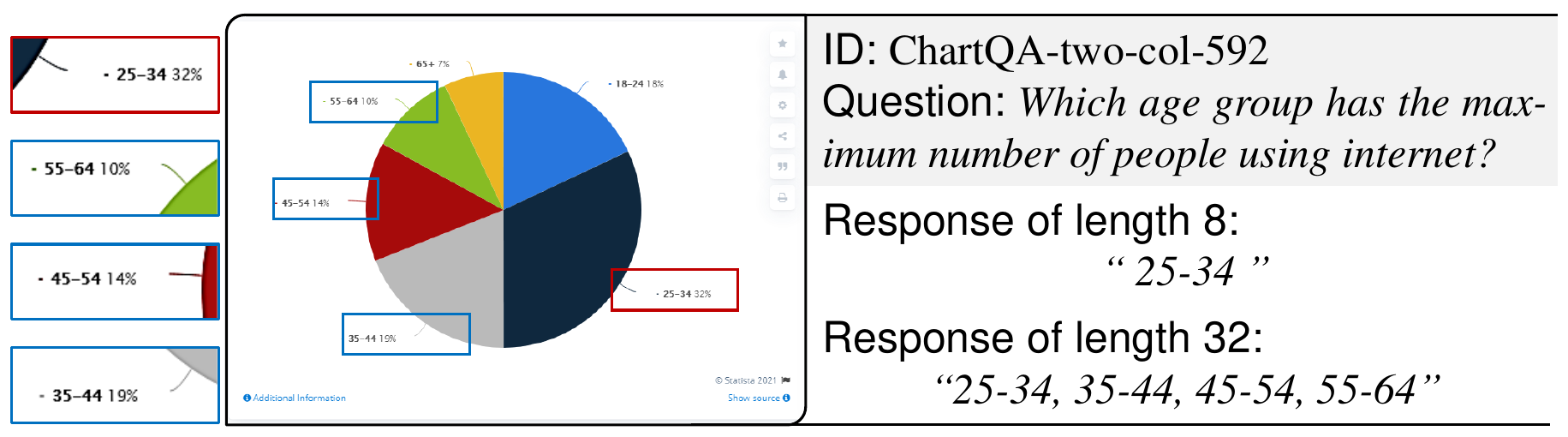}
    \caption{An example of Length Bias Phenomenon Collected from ChartQA. The response is generated by the model after 2-epoch diffusion tuning. The red and blue boxes on the left indicate the source locations of the numbers in the response.}
    \label{fig:length-bias-example}
\end{figure}
\paragraph{Length Bias}
Another notable weakness of models trained purely with diffusion is the presence of severe length bias, making them highly sensitive to the response length hyperparameter. This issue is especially pronounced in chart-based QA tasks. \cref{fig:auto-abl:2,fig:auto-abl:3} illustrate the relationship between response length and accuracy on AI2D and ChartQA.  The AT+DT model maintains relatively stable performance across varying lengths, whereas the pure diffusion model shows dramatic performance drops as length increases. For example, in ChartQA, accuracy drops from 42.7\% to 8.6\% as the response length increases from 8 to 32.
\cref{fig:length-bias-example} shows an example from ChartQA, where the pure diffusion model exhibits a tendency to translate all mask tokens into normal textual tokens. This behavior leads the model to continuing to generate text even after producing a correct answer, as it lacks the ability to ``stop'' appropriately.

\section{Sensitivity Analysis on Generation Parameters}

We conducted a sensitivity analysis on the following hyperparameters: Response Length ($L$), Number of Steps ($k$), Confidence Threshold ($\gamma$), Temperature ($t$), and Top-p ($p$). 
\textbf{1. Response Length ($L$)}  

Fixing $k=L$, $\gamma=1$, $t = 0$, and $p = 0.01$, we analyzed the sensitivity to response length. While response length does affect performance, its impact is generally minor across most datasets.

\begin{table}[h]
\centering
\begin{tabular}{ccccc}
\toprule
L  & CharQA & SQA  & POPE & AI2D \\ \midrule
16 & 63.3   & 76.7 & 86.3 & 73   \\
8  & 63.0   & 77.2 & 86.1 & 72.9 \\
4  & 56.4   & 77.5 & 86.2 & 72.9 \\ \bottomrule
\end{tabular}
\caption{Sensitivity analysis of $L$.}
\end{table}

\textbf{2. Steps ($k$)}  

Fixing $L = 8$, $\gamma = 1$, $t = 0$, and $p = 0.01$, we varied $k$ and observed that reducing $k$ degrades performance, indicating a trade-off between decoding speed and performance.

\begin{table}[h]
\centering
\begin{tabular}{ccccc}
\toprule
k & CharQA & SQA  & POPE & AI2D \\ \midrule
8 & 63.0   & 77.2 & 86.1 & 72.9 \\
4 & 57.2   & 76.1 & 85.9 & 72.8 \\
2 & 52.2   & 76.2 & 85.5 & 71.1 \\ \bottomrule
\end{tabular}
\caption{Sensitivity analysis of $k$.}
\end{table}

\textbf{3. Confident Decoding Threshold ($\gamma$)}  

Fixing $L = 8$, $t = 0$, and $p = 0.01$, we evaluated the effect of confident decoding. Within the range $\gamma \in [0.8, 1.0]$, confident decoding remains stable and does not significantly impact performance across most datasets.

\begin{table}[h]
\centering
\begin{tabular}{ccccc}
\toprule
$\gamma$ & CharQA & SQA  & POPE & AI2D \\ \midrule
1        & 63.0   & 77.2 & 86.1 & 72.9 \\
0.9      & 62.5   & 77.0 & 86.4 & 73.1 \\
0.8      & 61.9   & 77.1 & 86.3 & 73.1 \\
0.7      & 60.1   & 76.8 & 85.9 & 72.8 \\ \bottomrule
\end{tabular}
\caption{Sensitivity analysis of $\gamma$.}
\end{table}

\textbf{4. Temperature ($t$)}  

Fixing $L = k = 8$, $\gamma = 1$, and $p = 0.95$, we varied the temperature. The model performs best when $t < 0.2$; higher values of $t$ significantly harm performance.

\begin{table}[h]
\centering\begin{tabular}{ccccc}
\toprule
t   & CharQA & SQA  & POPE & AI2D \\ \midrule
2   & 48.5   & 57.4 & 77.9 & 55.6 \\
1   & 57.9   & 74.4 & 83.1 & 70.3 \\
0.2 & 63.0   & 76.9 & 86.3 & 73.4 \\
0   & 63.0   & 77.2 & 86.1 & 72.9 \\ \bottomrule
\end{tabular}
\caption{Sensitivity analysis of $t$.}
\end{table}

\textbf{5. Top P ($p$)}  

Fixing $L = k = 8$, $\gamma = 1$, and $t = 0.2$, we varied the top-p value and found that performance is largely unaffected. This suggests the model is not highly sensitive to $p$.

\begin{table}[h]
\centering
\begin{tabular}{ccccc}
\toprule
  $p$  & CharQA & SQA  & POPE & AI2D \\ \midrule
  0.95 & 63.6   & 76.9 & 86.2 & 73.4 \\
  0.5  & 63.0   & 77.2 & 86.3 & 72.7 \\
  0.1  & 62.9   & 77.2 & 86.1 & 73.0 \\ \bottomrule
\end{tabular}
\caption{Sensitivity analysis of $p$.}
\end{table}

\section{Further Discussion on Related Works}

\subsection{Multimodal Large Language Models}
Multimodal large language models (MLLMs) have rapidly progressed with several open-source series such as LLaVA, Eagle, InternVL, and Qwen-VL. LLaVA aligns vision encoders with LLMs via visual instruction tuning and demonstrates competitive performance using efficient training and external tool use \citep{llava,llava15,li2024llavanext-strong,liu2024llavanext}.
Eagle models explore vision-language integration through multiple encoders and long-context strategies, notably Automatic Degrade Sampling and Image Area Preservation \citep{eagle,eagle2,eagle-2.5}.
InternVL focuses on scaling vision-language models, evolving from large vision encoders to joint multimodal pretraining, and achieves state-of-the-art results with models like InternVL2.5 and InternVL3 \citep{internvl,intern2vl,internvl3,internvl-2.5}. 
Qwen-VL introduces multilingual vision-language models with dynamic resolution processing and bounding-box generation, achieving competitive performance across diverse tasks \citep{qwen-vl,qwen2-vl,qwen25vl}.

With the advancement of discrete large language models (dLLMs), discrete diffusion has also been applied to multimodal understanding tasks. Concurrent with Dimple, models such as UniDisc~\citep{unidisc}, MMaDA~\citep{mmada}, LLaDA-V~\citep{lladav}, and LaViDA~\citep{lavida} have emerged. Among them, UniDiscc~\citep{unidisc} and MMaDA~\citep{mmada} are unified discrete diffusion models that jointly model multimodal and text generation within the discrete diffusion framework. 
Dimple, LLaDA-V~\citep{lladav}, and LaViDA~\citep{lavida} are all multimodal understanding models. In terms of training, LLaDA-V~\citep{lladav} successfully scales the training dataset to 17 million samples. LaViDA~\citep{lavida} introduces complementary masking strategies for training data, while Dimple proposes a hybrid training method called ``First-Autoregressive-then-Diffusion''.
For inference, both Dimple and LaViDA~\citep{lavida} demonstrate the effectiveness of pre-filling techniques in discrete diffusion models. Additionally, Dimple introduces the Confident Decoding algorithm, the first method capable of dynamically controlling the number of generation steps in a discrete diffusion model, thereby balancing generation speed and quality.

\subsection{Parallel-Decoding}
Parallel decoding techniques have been widely used in image generation~\citep{maskgit,muse,mage,aig} and audio generation~\citep{impact,soudstorm,mag} to accelerate the generation process. There are two main approaches for selecting which token positions to decode. One approach relies on the magnitude of confidence score: these methods first predict tokens at all positions, compute confidence score for each position, and then retain the predictions at positions with high confidence scores ~\citep{maskgit,muse,mage,soudstorm,mag}. Another approach randomly selects a subset of positions to decode~\citep{aig,impact}.

In these works, the number of positions decoded  at each timestep is determined by a pre-defined scheduling function. For example, the commonly used cosine schedule imposes a fixed slow-to-fast generation pattern. A limitation of this strategy is that the top-k tokens with the highest confidence scores are always decoded, even if some of them have low absolute confidence. For example, if  and the top two confidence scores are 0.95 and 0.1, both tokens will be decoded, despite the second one being highly uncertain.

To address this limitation, our confident decoding algorithm differs from previous works by allowing the model to dynamically determine the number of tokens to decode at each timestep based on its own confidence score. Specifically, we set a confidence threshold and only decode tokens whose predicted confidence exceeds this threshold. This prevents the model from decoding uncertain tokens and allows it to adaptively adjust decoding speed depending on the input query. While both our approach and prior works make use of confidence scores, our method introduces greater flexibility by not fixing the number of tokens decoded per step.

\subsection{Diffusion Multimodal Large language Models}
In the main-text, we provide the discussion and performance comparison among dMLLMs. Here, we provide \cref{tab:compare_dmllm_tech} for a summarization of the differences among these models. Because UniDisc uses Held-out evaluation, it trains one model for each dataset. So we leave its \#Training Data as blank.
\begin{table}[t]
\makebox[\textwidth]{\resizebox{1\textwidth}{!}{
\begin{tabular}{@{}lcccccc@{}}
\toprule
                                                       & Dimple            & UniDisc                                         & MMaDA     & LLaDA-V   & LaViDA-L  & LaViDA-D  \\\midrule
Model Size                                             & 7B                & 1.4B                                            & 8B        & 8B        & 8B        & 7B        \\
\#Training Data                                        & 1.3M              & * & $\sim$2M  & 17M       & 1.6M      & 1.6M      \\
Training Loss                                          & AR then Diffusion & Diffusion                                       & Diffusion & Diffusion & Diffusion & Diffusion \\
Reinforcement   Leraning                               &                   &                                                 & \checkmark         &           &           &           \\
Complementary   Masking Training                       &                   &                                                 &           &           & \checkmark         & \checkmark         \\
Inference with Prefilling                              & \checkmark                 &                                                 &           &           & \checkmark         & \checkmark         \\
Confident Decoding with Dynamic Inference   Scheduling & \checkmark                 &                                                 &           &           &           &          \\\bottomrule
\end{tabular}}}
\caption{Technical comparison between Dimple and other approaches, highlighting its AR-then-diffusion training paradigm and the use of Prefilling and Confident Decoding for faster inference.
}
\vspace{-0.5em}
\label{tab:compare_dmllm_tech}
\end{table}

\section{Discussions}
\paragraph{Limitations.} Dimple-7B is trained on less than 1/20th of the data used by state-of-the-art models such as Qwen2.5-VL and Eagle2. Consequently, while Dimple demonstrates competitive performance among models of similar scale, it still falls short compared to the most powerful models. Therefore, scaling up training data and model parameters is a key future direction for dMLLM.
Although confident decoding improves efficiency, the full attention mechanism in discrete diffusion still incurs quadratic memory and computational complexity, limiting scalability to longer sequences and low-resource environments. Thus, further investigation into optimizing dMLLM inference is essential.

\paragraph{Broader Impact} Dimple's safety concerns arise from both the underlying DLM model and Dimple's vision encoding process. Like other MLLMs, it is also susceptible to issues such as hallucination and generation bias. 

\paragraph{Safeguards.} The following Usage Disclosure is included in the Dimple document. ``We release the Dimple model for academic and non-commercial research purposes only. Users are strongly discouraged from using the model to generate misleading or harmful content, including but not limited to disinformation, impersonation, or harassment. Dimple is not designed or validated for use in safety-critical domains such as medical decision-making, autonomous vehicles, or legal advice systems.
''

\section{Examples}
In this section, we provide additional examples of Dimple's generations to qualitatively showcase its capabilities. A typical phenomenon is that the existence of randomness in the decoding algorithm affects the length of the meaningful text generated. Here, meaningful text refers to tokens excluding special tokens such as padding token and mask token. Given a response length, shorter outputs indicate a higher proportion of meaningless padding tokens, whereas longer outputs imply fewer such tokens. Here, the decoding algorithm specifically refers to the method used to select which tokens are decoded, such as MaskGIT. The randomness in the decoding algorithm is affected by two factors: first, whether token selection is performed via sampling or by greedily choosing the position with the highest confidence; second, the value of the temperature parameter.
When there is no randomness, Dimple tends to generate shorter responses. In contrast, with randomness, Dimple is more likely to produce longer responses. 

Not all questions exhibited the aforementioned phenomenon.
Some questions are inherently unaffected. For instance, as shown in \cref{tab:exp_sup:1}, mathematical reasoning tasks naturally require multi-step reasoning, resulting in relatively long answers regardless of the presence of randomness. In contrast, in \cref{tab:exp_sup:2,tab:exp_sup:3}, tasks such as pure OCR recognition or counting—which involve only perception without the need for deeper understanding or analysis—tend to yield short answers, regardless of the decoding strategy.

However, for tasks where the response length is not intrinsically fixed, the presence or absence of randomness in decoding has a significant impact. For example, in \cref{tab:exp_sup:4,tab:exp_sup:5,tab:exp_sup:6}, tasks that require both perception and visual understanding, or involve describing images, clearly show that decoding with randomness tends to generate much longer responses.

In \cref{tab:exp_sup:4,tab:exp_sup:5,tab:exp_sup:6}, the response length is set to be 128 and the decodinng algorithm is MaskGIT. For the ``without randomness'' setup, the temperature of the MaskGIT is 0. For the ``with randomness'' setup, the temperature is 0.4.

\begin{table}[h]
\centering 
\renewcommand{\arraystretch}{1.3} 
\begin{tabular}{@{}lp{12cm}@{}}
\toprule
 \multicolumn{2}{l}{Mathematical Reasoning}\\ \midrule
 &  \includegraphics[width=0.3\linewidth]{fig/mmvet/v1_0.png}\\
 User:&  First please perform reasoning, and think step by step to provide best answer to the following question: What is x in the equation?\\
 Dimple:&  The equation in the image is (X+3)² = 4. To find the value of x, we need to solve the equation. First, we take the square root of both sides to get rid of the square. This gives us x + 3 = ±2. Then, we subtract 3 from both sides to isolate x. This gives us x = -3 ± 2. This results in two possible values for x: x = -3 + 2 = -1 and x = -3 - 2 = -5. So, the value of x is either -1 or -5.\\ 
 \bottomrule
\end{tabular}
\caption{Example of Dimple Generation. The image and question are collected from the MMVet dataset~\citep{mmvet}. This example demonstrates Dimple's capabilities in OCR and mathematical reasoning.  
}
\label{tab:exp_sup:1}
\end{table}
\begin{table}[h]
\centering 
\renewcommand{\arraystretch}{1.3} 
\begin{tabular}{@{}lp{12cm}@{}}
\toprule
 \multicolumn{2}{l}{OCR}\\ \midrule
 &  \includegraphics[width=0.3\linewidth]{fig/mmvet/v1_93.jpg}\\
 User:&  What are all the scene text in the image?\\
 Dimple:&  Connecticut law of 1642, If any man or woman be a witch- that is, hath or consulteth with a familiar spirit--they shall be put to death\\ 
 \bottomrule
\end{tabular}
\caption{Example of Dimple Generation. The image and question are collected from the MMVet dataset~\citep{mmvet}. This example demonstrates Dimple's capabilities in pure OCR. 
}
\label{tab:exp_sup:2}
\end{table}

\begin{table}[h]
\centering 
\renewcommand{\arraystretch}{1.3} 
\begin{tabular}{@{}lp{12cm}@{}}
\toprule
 \multicolumn{2}{l}{Counting}\\ \midrule
 &  \includegraphics[width=0.3\linewidth]{fig/mmvet/v1_86.jpg}\\
 User:&  How many tomatoes are there?\\
 Dimple:&  Five\\ \bottomrule
\end{tabular}
\caption{Example of Dimple Generation. The image and question are collected from the MMVet dataset~\citep{mmvet}. This example demonstrates Dimple's capabilities in counting. 
}
\label{tab:exp_sup:3}
\end{table}

\begin{table}[h]
\centering 
\renewcommand{\arraystretch}{1.3} 
\begin{tabular}{@{}lp{9cm}@{}}
\toprule
 \multicolumn{2}{l}{Chart Understanding}\\ \midrule
 &  \includegraphics[width=0.3\linewidth]{fig/mmvet/v1_52.png}\\
 User:&  What is the solution if the lamp is plugged in and the bulb has burned out?\\
 Dimple (w/o Randomness):&  The solution is to replace the bulb.\\ 
 Dimple (w Randomness):&  If the lamp is not plugged in and the bulb has burned out, the solution would be to replace the bulb. This is because if the bulb is dead and it's not plugged into the power source, then replacing the bulb is necessary to restore the functionality of the lamp.\\ 
 \bottomrule
\end{tabular}
\caption{Example of Dimple Generation. The image and question are collected from the MMVet dataset~\citep{mmvet}. This example demonstrates Dimple's capabilities in OCR and chart understanding. 
}
\label{tab:exp_sup:4}
\end{table}

\begin{table}[h]
\centering 
\renewcommand{\arraystretch}{1.3} 
\begin{tabular}{@{}lp{10cm}@{}}
\toprule
 \multicolumn{2}{l}{Recognition}\\ \midrule
 &  \includegraphics[width=0.3\linewidth]{fig/mmvet/v1_107.jpg}\\
 User:&  First please perform reasoning, and think step by step to provide best answer to the following question: Is this airplane taking off or landing?\\
 Dimple (w/o Randomness):&  The airplane is taking off.\\ 
 Dimple (w Randomness):&  The airplane is taking off because, as can be seen in the image, the landing gear is off the ground, and the lower part of the wing is visible. This indicates the airplane is not yet on the runway, and it is in the air.\\ \bottomrule
\end{tabular}
\caption{Example of Dimple Generation. The image and question are collected from the MMVet dataset~\citep{mmvet}. This example demonstrates Dimple's capabilities in recognition. 
}
\label{tab:exp_sup:5}
\end{table}

\begin{table}[h]
\centering 
\renewcommand{\arraystretch}{1.3} 
\begin{tabular}{@{}lp{10cm}@{}}
\toprule
 \multicolumn{2}{l}{Chicken Nugget Map Meme}\\ \midrule
 &  \includegraphics[width=0.3\linewidth]{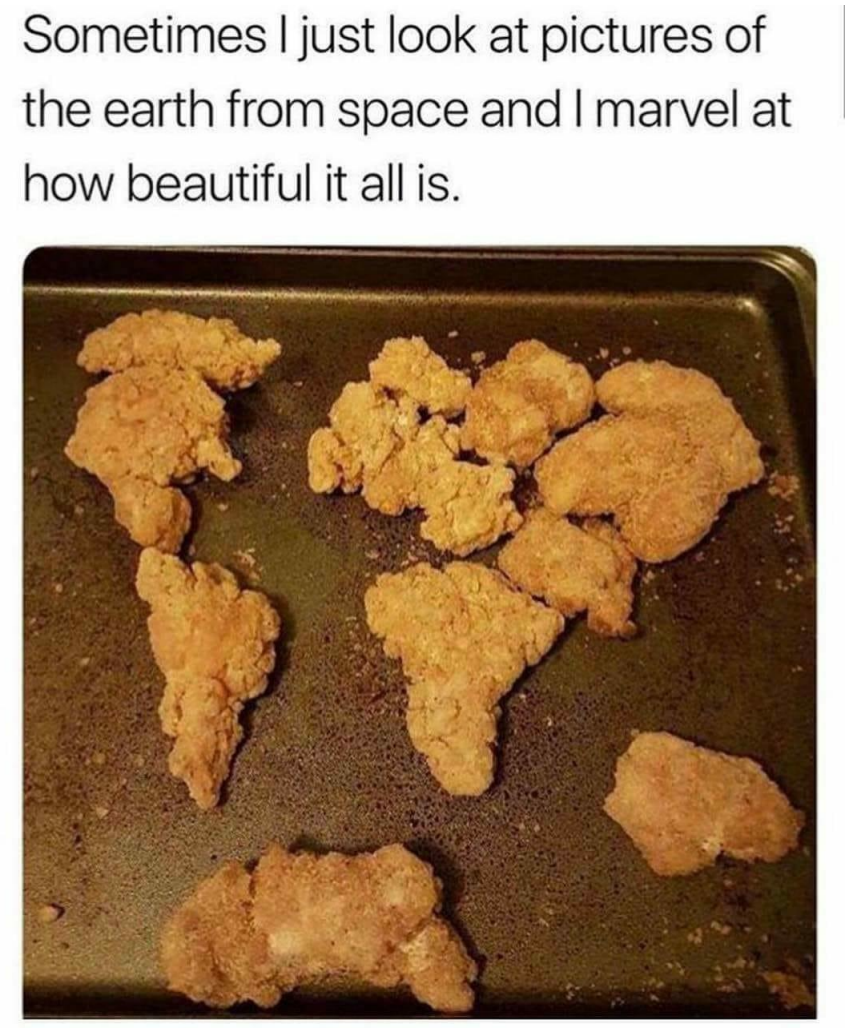}\\
 User:&  Can you explain this meme in detail?\\
 Dimple (w/o Randomness):&  The meme features a tray of golden chicken nuggets, arranged to resemble a world map.\\ 
 Dimple (w Randomness):&  The meme is a humorous image featuring a tray of golden breaded or fried food on a baking sheet, along with the text "Sometimes I just look at the pictures of the Earth from space and I marvel at how beautiful it is." The meme appears to be comparing the appearance of the food on the tray to a stunning view of a planet from space, emphasizing the beauty of the Earth in relation to the food. The contrast between the food on the tray and the celestial image adds to the comparison in a playful and entertaining way. The overall message of the meme is to highlight the wonder and beauty of the natural world.\\ \bottomrule
\end{tabular}
\caption{Example of Dimple Generation. The image and question are collected from the LLaVA Paper~\citep{llava}. This is a typical example demonstrates the model's ability to understand memes and the abstract humor embedded within them. 
}
\label{tab:exp_sup:6}
\end{table}

\begin{table}[h]
\centering 
\renewcommand{\arraystretch}{1.3} 
\begin{tabular}{@{}lp{13cm}@{}}
\toprule
 \multicolumn{2}{l}{GSM8K Question: Driect Instruction}\\ \midrule
 User:&  The cafeteria had 23 apples. If they used 20 to make lunch and bought 6 more, how many apples do they have? Directly answer the question.\\
 Dimple :&  9\\ \bottomrule
\end{tabular}
\caption{Example of Dimple Generation with explicit instruction. 
}
\label{tab:exp_sup:7}
\end{table}

\begin{table}[h]
\centering 
\renewcommand{\arraystretch}{1.3} 
\begin{tabular}{@{}lp{13cm}@{}}
\toprule
 \multicolumn{2}{l}{GSM8K Question: In-Context Learning}\\ \midrule
 User:&  Q: Roger has 5 tennis balls. He buys 2 more cans of tennis balls. Each can has 3 tennis balls. How many tennis balls does he have now?\\
 &
A: <think> Roger started with 5 balls. 2 cans of 3 tennis balls each is 6 tennis balls. 5 + 6 = 11. </think> The answer is 11.\\
&Q: The cafeteria had 23 apples. If they used 20 to make lunch and bought 6 more, how many apples do they have?
\\
 Dimple :&  <think> They started with 23 apples and used 20 so 23 - 20 = 3 apples. They bought 6 more so they have 3 + 6 = 9 apples. </think> The answer is 9.\\ \bottomrule
\end{tabular}
\caption{Example of Dimple Generation with In-Context Learning. 
}
\label{tab:exp_sup:9}
\vspace{6em}
\end{table}

\end{document}